\documentclass[sn-mathphys,Numbered]{sn-jnl}
\usepackage{graphicx}%
\usepackage{multirow}%
\usepackage{amsmath,amssymb,amsfonts}%
\usepackage{amsthm}%
\usepackage{mathrsfs}%
\usepackage[title]{appendix}%
\usepackage{xcolor}%
\usepackage{textcomp}%
\usepackage{manyfoot}%
\usepackage{booktabs}%
\usepackage{algorithm}%
\usepackage{algorithmicx}%
\usepackage{algpseudocode}%
\usepackage{listings}%
\usepackage{url}
\usepackage[marginal]{footmisc}

\theoremstyle{thmstyleone}%
\theoremstyle{thmstyletwo}%

\theoremstyle{thmstylethree}%

\usepackage{geometry}
\newgeometry{left = 3cm, right = 3cm, top = 3cm , bottom=3cm}
\begin{document}

\title[MaCo: A Radiography-Reports Foundation Model]{Enhancing Representation in Radiography-Reports Foundation Model: A Granular Alignment Algorithm Using Masked Contrastive Learning}

\author[1,2,3,6]{\fnm{Weijian} \sur{Huang}}
\author[1,6]{\fnm{Cheng} \sur{Li}}
\author[4]{\fnm{Hong-Yu} \sur{Zhou}}
\author[1,2,3]{\fnm{Hao} \sur{Yang}}
\author[1,2,3]{\fnm{Jiarun} \sur{Liu}}
\author[2]{\fnm{Yong} \sur{Liang}}
\author[1]{\fnm{Hairong} \sur{Zheng}}
\author[5]{\fnm{Shaoting} \sur{Zhang}}

\author *[1]{\fnm{Shanshan} \sur{Wang}} \email{ss.wang@siat.ac.cn}

\affil[1]{\orgname{Paul C. Lauterbur Research Center for Biomedical Imaging, Shenzhen Institute of Advanced Technology}, \orgaddress{\city{Shenzhen}, \country{China}}}
\affil[2]{\orgname{Pengcheng Laboratory}, \orgaddress{\city{Shenzhen}, \country{China}}}
\affil[3]{\orgname{University of Chinese Academy of Sciences}, \orgaddress{\city{Beijing}, \country{China}}}
\affil[4]{\orgname{Department of Biomedical Informatics, Harvard University}, \orgaddress{\country{United States}}}
\affil[5]{\orgname{Shanghai AI Lab}, \orgaddress{\city{Shanghai}, \country{China}}}

\abstract{Recently, multi-modal vision-language foundation models have gained significant attention in the medical field. While these models offer great opportunities, they still face crucial challenges, such as the requirement for fine-grained knowledge understanding in computer-aided diagnosis and the capability of utilizing very limited or even no task-specific labeled data in real-world clinical applications. In this study, we present MaCo, a masked contrastive chest X-ray foundation model that tackles these challenges. MaCo explores masked contrastive learning to simultaneously achieve fine-grained image understanding and zero-shot learning for a variety of medical imaging tasks. It designs a correlation weighting mechanism to adjust the correlation between masked chest X-ray image patches and their corresponding reports, thereby enhancing the model's representation learning capabilities. To evaluate the performance of MaCo, we conducted extensive experiments using 6 well-known open-source X-ray datasets. The experimental results demonstrate the superiority of MaCo over 10 state-of-the-art approaches across tasks such as classification, segmentation, detection, and phrase grounding. These findings highlight the significant potential of MaCo in advancing a wide range of medical image analysis tasks.}

\keywords{Multi-Modal Representation Learning, Vision-Language Representation Learning, Foundation Model}

\maketitle
\renewcommand{\thefootnote}{}
\footnote{$^6$ These authors contributed equally to this work.}

\section*{Introduction}\label{sec1}
Recent advances in machine learning have revolutionized the potential of automated diagnostic systems (ADS) by achieving expert-level performance, making it feasible to use deep learning to improve the clinical workflow \cite{rajpurkar2023current, chang2023mining, liu2024swin}. These ADS have demonstrated their efficacy in addressing various routine clinical tasks, such as disease diagnosis and lesion quantification, through training diverse machine learning models \cite{rajpurkar2023current}. However, this traditional approach of training separate models from scratch for specific applications has inherent limitations. It is computationally expensive and demands a considerable amount of manually annotated data, which fundamentally limits the development and scalability of medical applications \cite{acosta2022multimodal, moor2023foundation}. As a result, there is an urgent need to explore alternative approaches that can improve the effectiveness of ADS while mitigating these challenges \cite{wu2023medklip}.

One promising solution is to develop medical foundation models that can handle multiple clinical applications simultaneously and leverage pre-trained models to reduce the dependency on large annotated datasets \cite{tiu2022expert, isbiliu, wu2023medklip, moor2023foundation, zhou2023foundation, isbiyang, zhou2023unified}. These models can be trained on diverse and representative image-based datasets using self-supervised methods that do not require annotations, allowing them to learn robust and transferable feature representations that can be used across various tasks and domains \cite{zhou2022generalized, isbihuang}. By incorporating simple task-based heads with the well-learned feature representations from the foundation model, these methods can achieve good performance in specific tasks without the need for extensive manual annotations \cite{9879206}. This reduces the labeling burden on clinical experts and enhances the potential for clinical deployment. 

However, with the expanding adoption of these methods, researchers are facing increasing challenges \cite{sutton2020overview}. These challenges predominantly stem from the need for high performance in clinical deployment settings. Integrating expert knowledge with ADS has demonstrated promising results, as it combines human insight with data-driven machine learning approaches \cite{wu2023medklip, zhang2023knowledge, zhou2023transformer}. This approach holds the potential to generate more reliable and intuitive results, making it a valuable tool for improving the performance of ADS \cite{acosta2022multimodal}. Coincidentally, radiology reports obtained from daily clinical examinations often contain valuable information regarding the healthcare history, imaging manifestations, and disease severity of the patients. These reports can serve as a valuable source of human knowledge, which can be leveraged to augment the capabilities of ADS. However, extracting meaningful information from radiology reports remains a pressing issue due to their highly subjective and unstructured nature, which can vary depending on the individual style of the clinical physician. Effective integration of rich human knowledge from radiology reports with machine learning models continues to be an ongoing challenge.

Many endeavors have been made to leverage expert knowledge from clinical reports \cite{zhou2022generalized, huang2023visual}. These efforts can be broadly categorized into two branches. The first branch focuses on improving radiological representations for downstream tasks through fine-tuning. These methods employ sophisticated self-supervised pretext tasks, such as masked autoencoders (MAE) \cite{9879206} or combining with high-resolution reconstruction (HR) \cite{zhou2022advancing}, to obtain robust image representations. These representations are then integrated with the textual information to enhance the performance of downstream fine-tuning tasks \cite{zhou2022advancing, chen2022multi}. The second branch draws inspiration from contrastive learning approaches \cite{radford2021learning} and aims to align the distributions of image features and text features \cite{wu2023medklip, 9710099, boecking2022making}. These methods not only achieve comparable fine-tuning performance but also possess zero-shot capabilities to cope with the complex and diverse clinical environment. We propose that striking a proper balance between these methods would be advantageous. However, such attempts have not been extensively explored in the medical field thus far.

In this paper, we focus on two key aspects of building a vision-language foundation model for chest X-ray analysis. Firstly, we emphasize the significance of incorporating clinical reports to enhance the model's semantic comprehension of radiographic images \cite{zhou2022advancing, wu2023medklip, zhang2023knowledge}. We believe that integrating clinical reports, which contain rich professional knowledge, into image-based models is a crucial advancement in the realm of precision medicine. Secondly, we advocate for the foundation model to possess a certain level of capability even in extreme scenarios with limited annotations \cite{rajpurkar2023current}, where only a scarcity of labeled data may exist for downstream tasks. This ensures enhanced applicability of the constructed foundation model, even in situations where no annotations are available for specific tasks. To address these requirements, we introduce a masked contrastive chest X-ray foundation model (MaCo), which is designed to facilitate cross-modal vision-language knowledge comprehension, thereby enhancing feature representation learning. As depicted in Fig. 1 (a), MaCo integrates the strengths of pretext task-based learning and contrastive learning, while incorporating a correlation weighting mechanism to further enhance the capabilities of representation learning. Through extensive experiments, we have thoroughly evaluated the effectiveness of MaCo in various downstream fine-tuning as well as zero-shot learning tasks. Experimental results demonstrate the superiority of MaCo over 10 existing state-of-the-art models. The exceptional performance achieved by MaCo in zero-shot learning tasks highlights its potential to reduce annotation costs in medical applications.

\section*{Results}\label{sec2}
To validate the effectiveness of MaCo as a foundational model for chest X-ray analysis, we begin by evaluating its performance on various fine-tuning tasks, including classification, segmentation, and detection tasks while utilizing different numbers of annotated fine-tuning data. Then, we provide qualitative and quantitative results to showcase MaCo's zero-shot phrase-grounding and zero-shot classification capabilities. Finally, visualizations of the proposed weighting mechanism are presented to demonstrate how our network progressively targets disease-relevant regions. It should be noted that all the metrics for the comparison algorithms are directly drawn from their own publication reports. If they haven't reported the results for specific tasks, we follow the report results from \cite{zhou2022advancing, wu2023medklip, lovt}, unless otherwise specified.

\subsection*{Fine-tuning classification}
We present the fine-tuning results of various methods on classification tasks using three datasets, CheXpert, RSNA, and NIH ChestX-ray. Different ratios of annotated samples for fine-tuning were experimented with, and the results obtained by our proposed MaCo are compared with those generated by the currently prevailing pre-text-based non-contrastive learning methods and contrastive learning methods. While non-contrastive learning methods lack zero-shot capabilities, which may limit their applicability in clinical settings, we include these methods in our comparative analyses to achieve a more comprehensive evaluation.

We conduct comparative analyses between MaCo and four state-of-the-art non-contrastive learning methods, Ark, M3AE, REFERS, and MRM, and five state-of-the-art contrastive learning methods, ConVIRT, GloRIA, BioViL, MedKLIP, and M-FLAG. The results are presented in Table \ref{ftcls}. MRM adopts both masked autoencoder and high-resolution reconstruction as the pretext tasks. It achieves promising fine-tuning classification performance under different settings, surpassing the five existing state-of-the-art contrastive learning methods. However, it should be noted that MRM, as well as the other non-contrastive learning methods, sacrifices the zero-shot capabilities. In addition, they cannot perform zero-shot phrase grounding for text-image correlation visualization, which is an important strategy to enhance the model's explainability. This trade-off may potentially reduce their scalability and applicability in real-world clinical applications. Besides, the performance advantage of non-contrastive learning methods over contrastive learning methods may diminish when the scale of pre-training datasets in the medical domain increases, as it has been shown that contrastive learning methods can benefit more from larger datasets \cite{radford2021learning}. In the current setting, the models were pre-trained with MIMIC-CXR, which comprises 200,000 image-report pairs. This dataset size is considerably smaller compared to natural datasets, which can exceed 400 million samples (CLIP \cite{radford2021learning}). Nevertheless, MaCo achieves a classification performance comparable to MRM while retaining the capabilities of zero-shot learning and text-image correlation visualization. Compared to the five existing contrastive learning methods, MaCo achieves the highest scores across different datasets and different ratios of utilized fine-tuning labeled data.

The results of various methods on disease-level classification using the NIH ChestX-ray dataset are presented in Table \ref{nihftcls}. All methods were fine-tuned using 100\% annotated data. To provide a more comprehensive evaluation, we introduce four additional image-based pretext task comparative methods, namely Model Genesis \cite{zhou2021models}, C2L \cite{zhou2020comparing}, Context Restoration \cite{chen2019self}, and TransVW \cite{haghighi2021transferable}. Consistent with our observations in dataset-level classification tasks, among the different non-contrastive learning methods, MRM demonstrates the best results in this disease-level classification task. When it comes to contrastive learning methods with zero-shot capabilities, MedKLIP and our MaCo show promising performance. Leveraging the rich information embedded in medical reports, MedKLIP demonstrates advanced results in classifying certain diseases, achieving an AUC score of 82.8$\%$ for consolidation, 90.8$\%$ for edema, and 98.0$\%$ for hernia. On the other hand, our proposed MaCo excels in achieving superior performance across 11 other disease categories when compared to other contrastive learning methods (ConVIRT, GLoRIA, BioViL, and MedKLIP).

Additionally, the GFLOPS (Giga Floating Point Operations Per Second) of our MaCo and different existing methods (MRM, GLoRIA, and BioViL) were measured using the open-source package 'thop' to evaluate the computational complexity. The GFLOPS for MRM, GLoRIA, and BioViL were recorded as 12.7, 16.2, and 19.4, respectively. Our proposed algorithm, MaCo, demonstrated a GFLOPS value of 16.9. These values indicate that MaCo's computational resource consumption is within a reasonable range.

Overall, in this downstream fine-tuning classification task, MaCo has demonstrated comparable performance compared with non-contrastive learning methods that lack zero-shot capabilities. Furthermore, when compared to methods with zero-shot capabilities, MaCo outperforms them with substantial margins in terms of classification performance. These observations validate the effectiveness of MaCo for this specific task, making it a highly promising method for relevant clinical applications.

\subsection*{Fine-tuning segmentation}
In this section, we discuss the segmentation results obtained by different methods through fine-tuning with 10$\%$ and 100$\%$ annotated data. We conducted experiments on two datasets, SIIM and COVID Rural, and compared our MaCo with eight state-of-the-art methods. Results are provided in Table \ref{ftseg}.

Our MaCo consistently outperforms the eight state-of-the-art approaches in all experimental settings. Specifically, on the SIIM dataset, when the annotated fine-tuning sample ratio is set to 10$\%$, MaCo achieves slightly better performance with a Dice score of 72.6$\%$ than the second-best method, MedKLIP, which achieves a Dice score of 72.1$\%$. As the annotated sample ratio increases to 100$\%$, MaCo demonstrates significant improvement, increasing the Dice score by 10\% compared to MedKLIP. This highlights MaCo's ability to capitalize on additional labeled data to enhance its feature representation and segmentation accuracy. On the COVID Rural dataset, MaCo's performance enhancement is even more impressive, surpassing the eight comparative approaches by significant margins. Under both annotated sample ratios, MaCo achieves Dice scores that are more than 30\% higher than the best comparative method, MedKLIP.

These experiments highlight the advantages of MaCo in terms of segmentation performance, showcasing its potential in reducing reliance on manual labeling and improving the efficiency of chest X-ray image segmentation.

\subsection*{Fine-tuning detection}

We evaluated the performance of MaCo on the RSNA dataset for the task of object detection. It is worth noting that the majority of the contemporary state-of-the-art detection methods adopt feature pyramids to enhance detection performance \cite{li2022exploring}. However, there is a lack of robust extension methods that can obtain feature pyramids for models pre-trained on the Vision Transformer (VIT) backbone \cite{li2022exploring}. This limitation potentially hinders the advantages of VIT-based models in detection tasks. Notably, our proposed MaCo utilizes the VIT architecture as its image encoder. To solve this issue, we employed the detection framework of VITDET, which is one of the few methods capable of accommodating pre-trained VIT models for the detection task. Besides, we also implemented CLIP* based on the VIT backbone as a baseline for fair comparison.

Six state-of-the-art detection methods that use ResNet backbone for the extraction of feature pyramids are introduced for comparison (Table \ref{ftdetection}) \cite{he2016deep}. The results highlight the following observations: Firstly, among the ResNet-based approaches, LoVT achieves the best performance with a mean average precision (mAP) score of 13.2 at the annotated sample ratio of 10$\%$. However, at the annotated sample ratio of 100$\%$, CLIP outperforms LoVT. Secondly, as expected, there is a decline in performance when replacing the backbone of CLIP with VIT. Specifically, CLIP* experienced a 2.5 mAP score decrease at the annotated sample ratio of 100$\%$. Thirdly, compared to CLIP*, our proposed MaCo demonstrates superior detection performance under both annotated sample ratios. This indicates the effectiveness of MaCo in the task of object detection. Nevertheless, our findings suggest that while VIT-based approaches, including our proposed MaCo, show promise in object detection, further research is needed to develop effective methods for incorporating feature pyramids into VIT-based models. This endeavor is crucial to enhance their effectiveness and bridge the performance gap between ResNet-based and VIT-based models in object detections.

\subsection*{Zero-shot classification}
Zero-shot classification has recently emerged as an important task attracting significant attention in the field. It plays a crucial role in validating the performance of multi-modal models and addressing extreme annotation limitations in clinical environments \cite{tiu2022expert, wu2023medklip, zhang2023knowledge}. In this work, we conducted zero-shot experiments on three open-source datasets: NIH, RSNA, and SIIM. We compared the performance of our proposed MaCo with five state-of-the-art algorithms: ConVIRT, GloRIA, BioViL, MedKLIP, and CheXzero.

As depicted in Table \ref{zscls}, the results demonstrate that our proposed MaCo outperforms all five comparative algorithms across all three datasets. This indicates that MaCo is capable of better aligning the image feature space and text feature space, leading to improved zero-shot classification performance. Moreover, on the NIH dataset, we provide zero-shot classification performance of various methods across 14 disease categories, as shown in Supplementary Fig. 1. Overall, MaCo achieves the best classification performance in the majority of diseases.

These zero-shot classification experiments further validate the effectiveness of MaCo in reducing the reliance on manual annotations. This positions MaCo as a valuable tool in a wider range of clinical applications, where annotated samples are difficult and expensive to collect.

\subsection*{Zero-shot phrase grounding}
\label{secpg}
Interpretable visualization of the correlations between image and text modalities is necessary to establish clinical trust and remove barriers to clinical application. Phrase grounding serves as an effective tool to achieve this purpose. Here, we evaluate the zero-shot phrase-grounding performance of MaCo on the MS-CXR dataset, which provides description phrases and corresponding bounding boxes.

Thanks to the proposed correlation weighting mechanism, we were able to utilize the weights of the fully connected (FC) layer to perform phrase grounding (the FC layer used to generate the importance score shown in Fig. 1 (b)(ii)). Specifically, each weight in this FC layer corresponds to the importance of one image patch, and thus, it can be utilized for phrase-grounding evaluation. We first applied a softmax function (with a soft threshold $\tau^{w}$) to the weights ({$w = \{w_i\} \in \mathbb{R}^{N \times 1}$, $i=1, 2, ..., N$, and $N=196$ is the total number of image patches) of this FC layer, obtaining the normalized weights $\widehat{w} \in \mathbb{R}^{N \times 1}$. 
$\widehat{w}$ is then utilized to multiply with the patch-based image representations obtained from the image encoder $v_{enc} \in \mathbb{R}^{N\times C}$ ($C$ is the feature dimension), generating $\overline{w} \in \mathbb{R}^{N\times C}$. After that, $\overline{w}$ is multiplied with the text representations from the text encoder $t_{enc} \in \mathbb{R}^{1\times C}$ to generate a phrase-grounding score map $s_{pg} = \overline{w} \times t_{enc}^T \in \mathbb{R}^{N \times 1}$. Finally, $s_{pg}$ is reshaped and bilinearly upsampled to the dimension of the input image, which is then compared with the ground truth to calculate the contrast-to-noise ratio (CNR) and mean Intersection over Union (mIoU) scores for the characterization of phrase-grounding performance.

In Table \ref{pg}, we present the quantitative phrase-grounding results of MaCo as well as three existing methods: ConVIRT, GLoRIA, and BioViL. Among the three comparative methods, BioViL obtains the best metrics with a mIoU of 0.266 and a CNR of 1.027. However, BioViL is pre-trained using three datasets, whereas ConVIRT and GloRIA were pre-trained on only one dataset. Compared with the two methods pre-trained on the same dataset, ConVIRT and GLoRIA, MaCo achieves better performance in terms of both CNR and mIoU. Specifically, MaCo achieves a CNR of 1.144, even surpassing BioViL. The observed superior performance of MaCo can be attributed to the proposed correlation weighting mechanism, which synergistically combines contrastive learning and masked autoencoder (MAE) in a cohesive manner.

Qualitative phrase-grounding results are presented in Fig. 2. Visual-textual correlation heatmaps obtained by different methods for six instances are plotted. These examples encompass various diseases, including atelectasis, opacity, and cardiomegaly. Overall, compared to the two existing methods, GLoRIA and BioViL, MaCo generates stronger responses in disease regions that correspond to the phrases across different diseases, indicating its enhanced capability in capturing visual-textual correlations.

The above results demonstrate the effectiveness of MaCo in zero-shot phrase grounding. It achieves promising results both quantitatively and qualitatively.These results collectively emphasize the potential of MaCo as a powerful tool for interpreting multi-modal medical data.

\subsection*{Granular alignment analysis of the proposed correlation weighting mechanism}

To verify the effectiveness of the proposed correlation weighting mechanism in achieving granular alignment, we visualize the weights of the FC layer in Fig. 1 (b)(ii), which corresponds to the importance of each image patch. Following our demonstration in Sec. \ref{secpg}, we reshape the normalized weight $\widehat{w} \in \mathbb{R}^{N \times 1}$ to generate the weight map with the dimension of $\sqrt{N} \times \sqrt{N}$ and plot the weight map in Fig. 3. Each pixel in the weight map corresponds to the weight assigned to an image patch. During the initial training stage, the weights are dispersed without prominent positions, indicating that the network has yet to learn the distinctions between different patches. As the training progresses over epochs, the weights in the central region of the image (typically representing the lungs) gradually increase, while the weights in the background regions decrease. This shift indicates that the model focuses more on image patches near the lungs, considering them to contain more informative content compared to the background regions. The weight map not only visualizes the model's attention on different image patches but also holds the potential to enhance downstream task performance, as demonstrated in the following analysis.

In Supplementary Table 1, we list the phrase-grounding results with different $\tau^{w}$ values utilized in the softmax function (please refer to Sec. \ref{secpg} for details). We observed that the grounding performance changes with different $\tau^{w}$ settings. Specifically, when $\tau^{w}$ is set to 0.01, MaCo attains the highest CNR of 1.149, whereas with $\tau^{w}$ set to 0.02, MaCo achieves the highest mIoU of 25.5. As $\tau^{w}$ continues to decrease, the scores of CNR and mIoU begin to decrease. Considering both metrics, we selected $\tau^{w}=0.02$ for our final phrase-grounding evaluation in Sec. \ref{secpg}.

\subsection*{Ablation study}
In this section, we investigate the contributions of each component in MaCo through phrase-grounding and fine-tuning classification tasks, as shown in Table \ref{ablation}. We use CNR, mIoU, and pointing game (PG) scores to characterize the phrase-grounding results and AUC to characterize the classification results.

We start with the MAE model trained solely on the image modality as our baseline. This baseline model lacks the capability to leverage information from radiology reports, thus it cannot perform phrase grounding. For fine-tuning classification on the RSNA dataset, MAE achieves the lowest AUC scores of 83.2$\%$, 89.2$\%$, and 91.0$\%$ at the annotated sample ratios of 1$\%$, 10$\%$, and 100$\%$, respectively. The incorporation of a high-resolution reconstruction task(+HR) in Table \ref{ablation}, slightly enhanced the classification performance when compared to MAE. The introduction of CLIP (+CLIP) empowered the model with zero-shot capabilities, achieving a mIoU of 21.2$\%$ and a CNR of 0.860 in the zero-shot phrase-grounding task using the MS-CXR dataset. The introduction of CLIP also led to substantial improvements in the classification performance, underscoring the value of integrating expert knowledge from medical reports into the image representation learning model. Our final model that integrates all these components with a correlation weighting mechanism (+Correlation Weighting) achieves the best performance in both the zero-shot phrase-grounding and fine-tuning classification tasks. Specifically, the phrase-grounding scores in terms of CNR and mIoU increase significantly from 0.860 to 1.144 and from 21.2$\%$ to 25.5$\%$, respectively. Concurrently, the AUC scores for the fine-tuning classification task are improved from 90.9$\%$, 92.$\%$, and 93.0$\%$ to 91.5$\%$, 92.7$\%$, and 93.6$\%$ at the annotated sample ratios of 1$\%$, 10$\%$, and 100$\%$.

\section*{Discussion}\label{sec13}
Fine-grained knowledge understanding and fine-tuning with limited annotated data for downstream tasks pose significant challenges in the development of medical foundation models. In this paper, we propose MaCo, a approach that addresses these challenges by achieving granular alignment between radiography and reports and extracting fine-grained representations.

A comprehensive evaluation of the effectiveness of MaCo was conducted utilizing 6 open-source datasets, involving a range of label-efficient fine-tuning tasks such as classification, segmentation, and detection. More than 10 state-of-the-art methods were included in the comparative analysis. The results revealed that our proposed MaCo demonstrated promising prospects across a range of tasks. Additionally, we validated the zero-shot capabilities of MaCo through zero-shot classification and phrase-grounding tasks. Both qualitative and quantitative indicators showcased the superiority of MaCo compared to over the ten methods. Furthermore, we quantified the degree of correlation between the location of each radiograph patch and its corresponding report through the proposed correlation weighting mechanism. This analysis highlighted the model's capability in effectively discriminating regions of radiographs that the model tends to focus on, enhancing the reliability and acceptability of the model in clinical applications.

While MaCo has demonstrated promising performance as a chest X-ray foundation model, it still faces several limitations. Firstly, to further enhance the robustness and generalizability of MaCo, there is a need to increase the scale of MaCo's pre-training data for wider use. This can be achieved by collecting a more diverse range of medical images from various imaging equipment and incorporating reports from a larger number of clinical physicians. By expanding and diversifying the dataset, MaCo can learn from a broader range of cases, leading to improved performance on diverse real-world scenarios. Secondly, MaCo currently employs the widely used language model BERT for text encoding. However, with the proliferation of larger and more specialized language models, future attempts should consider utilizing larger and more clinically oriented language models to achieve more effective domain-specific language understanding. Lastly, challenges associated with clinical deployment, including data privacy concerns and ethical considerations, need to be investigated in the future.

In conclusion, this paper introduces MaCo, a chest X-ray foundation model designed to address the challenges of fine-grained knowledge understanding and limited annotation learning in the medical domain. MaCo incorporates granular alignment, leveraging the advantages of both pretext task learning and contrastive learning. The promising results obtained from fine-tuning and zero-shot generalization experiments underscore the potential of MaCo in advancing medical foundation models. This work opens up avenues for further research and development in the field, bringing us towards more effective and generalizable medical AI solutions.

\section*{Methods}\label{sec11}
The high cost of annotation has long been a persistent challenge in the medical field. One prevalent approach to alleviating the annotation reliance in downstream tasks is the utilization of pre-trained models. With the rapid advancements in natural language processing models in recent years, there has been a growing interest in integrating expert knowledge from clinical reports with medical images. In the following sections, relevant studies in the medical domain, specifically within the realm of self-supervised pretext task-based and contrastive learning models, will be introduced. These studies serve as the foundation for our proposed MaCo. We declare that the proposed methods comply with all relevant ethical regulations and have been approved for research by the Shenzhen Institute of Advanced Technology.

\subsection*{Pretext task-based methods}
The goal of pretext task-based methods is to learn semantically meaningful image representations without utilizing any downstream task annotations \cite{misra2020self, albelwi2022survey}. These pretext tasks typically involve self-supervised learning techniques, such as using randomly augmented images or training on down-sampled images for high-resolution reconstruction. One widely utilized pretext task-based method is MAE. MAE \cite{9879206} applies a random masking technique to image patches within the input data. Subsequently, a reconstruction decoder is employed to recover the masked regions. By engaging in the reconstruction process, MAE is able to learn image features that can be subsequently utilized for various downstream tasks. Due to its simplicity and effectiveness, MAE has gained considerable popularity, including in the domain of medical image-text modeling. Drawing inspiration from MAE, Zhou et al. \cite{zhou2022advancing} employed a similar masking mechanism in both the text branch and the image branch of their model (MRM). They leveraged the vision representation as a supplementary component to the text branch and enhanced the feature representations through back-propagation optimization. Similar to MRM, Chen et al. \cite{geng2022multimodal} also employed masking in both the image and text modalities with a single transformer to integrate and couple the features of the image and text modalities (M3AE).

Although the aforementioned methods have shown promising performance in downstream fine-tuning tasks, their zero-shot capabilities are constrained by the adopted modality coupling strategy. This limitation impede their ability to generalize to unseen tasks, especially when dealing with unlabeled datasets.

\subsection*{Contrastive learning-based methods}
Contrastive learning-based methods, on the other hand, have recently gained significant attention due to their unique zero-shot capabilities \cite{radford2021learning, jaiswal2020survey}. Contrastive learning aims to minimize the similarity distance between paired data points within a training batch while simultaneously maximizing the dissimilarity between unpaired data points. By leveraging this approach, the trained models become proficient in differentiating between paired and unpaired images and texts, thereby acquiring the ability to generalize to unseen data samples, known as zero-shot capabilities \cite{zhang2022contrastive}.

Zhang et al. \cite{zhang2022contrastive} were pioneers in introducing contrastive learning as a proxy task in the field of medicine. Their study demonstrated the efficacy of contrastive learning within the medical domain. Building upon this foundation, Wang et al. \cite{wang2022medclip} further investigated the impact of false negative samples on the performance of contrastive learning methods. Boecking et al. \cite{boecking2022making} recognized the distinct language patterns found in medical reports, prompting a redesign of the language model for medical vision-language processing. Bannur et al. \cite{bannur2023learning} and Zhou et al. \cite{zhou2022generalized} employed past radiology images and multi-view images, respectively, for joint training purposes. In more recent developments, Wu et al. \cite{wu2023medklip} and Zhang et al. \cite{zhang2023knowledge} integrated a report filter to extract medical entities and employed a more complex modal fusion module to aggregate features, thereby achieving improved results. On the other hand, to establish fine-grained correspondence between images and reports, Li et al. \cite{li2023unify} aligned visual and textual semantics at different levels with explicit constraints. Huang et al. \cite{9710099} proposed a local fine-grained weighting mechanism. This mechanism calculates the similarity between each word and image patches, resulting in word-level responses. Similarly, Wang et al. \cite{wang2022multi} introduced the concept of multi-granularity alignment to explicitly learn the correspondence between fine-grained vision and text tokens. 

These contrastive learning-based methods have achieved comparable performance in downstream fine-tuning tasks to those pretext task-based methods. More importantly, some methods, such as BioViL and GLoRIA, have demonstrated inspiring zero-shot capabilities, which greatly enhance the task generalization capability of medical models. 

\subsection*{MaCo}
We introduce MaCo, a chest X-ray radiography-report foundation model with zero-shot capabilities, based on masked contrastive learning. The motivation behind MaCo is to leverage the advantages of both contrastive learning-based and pretext task-based methods to acquire enhanced semantic latent representations. MaCo investigates the masked autoencoder along with contrastive learning to facilitate learning from paired radiological images and medical reports. Additionally, we propose a correlation weighting mechanism in MaCo that weights the contrastive loss based on the importance of sampled image patches. This mechanism helps prioritize informative patches, resulting in more effective learning and better representation of relevant features. Fig. 1 shows the framework of MaCo, which integrates the strengths of contrastive learning and pretext task methods. The detailed methodology will be introduced in the subsequent sections.

\subsubsection*{Masked high-resolution image reconstruction for image feature extraction}
To extract meaningful feature representations from input images, we adopt MAE proposed by He et al. \cite{9879206} as our primary image representation extractor. MAE employs a reconstruction pretext task that is elaborately designed to restore the masked image, thereby extracting meaningful representations of the image. 

Specifically, the input image is partitioned into regular, non-overlapping patches, and a subset of the patches is randomly sampled as the inputs of the model while the remaining ones are excluded. Let us define $B$ as the batch size, and $C$ as the feature dimension. $N=N^{s}+N^{msk}$ represents the total number of divided image patches, where $N^{s}$ and $N^{msk}$ correspond to the number of sampled and masked patches, respectively. The encoder's prediction, given the masked image as input, is represented by $v_{enc}$ with the size of $ B\times N^sC$,  and the decoder's prediction is represented by $v_{recon}$ with the size of $B\times NC$. Let $g_{recon}$ denote the corresponding ground truth that is partitioned in the same way as the input image. The loss function of the masked autoencoder reconstruction in a batch can be written as:

\begin{equation}
\centering
\label{Eq.1}
\mathcal{L}_{mae}=||v_{recon}^{msk}-g_{recon}^{msk}||^{2}
\end{equation}
where $||\cdot||$ represents the $L_2$ norm. Here, we only focus on the recovery of the masked patches, such that $v_{recon}^{msk}$ and $g_{recon}^{msk}$ are the recovery of the masked patches and its corresponding ground-truth patches.

High-resolution reconstruction is also an effective pre-training approach in capturing the detailed representations of images \cite{zhou2022advancing}. This method takes low-resolution images as inputs for the image encoder and imposes constraints on the image decoder using original high-resolution images.

In MaCo, we incorporate both masked image reconstruction and high-resolution reconstruction as pre-text tasks during pre-training. The input image is firstly down-sampled to a smaller resolution. In this work, the down-sampling ratio is set to $2$. Then, following the practice adopted in MAE, the down-sampled input image is partitioned into $N$ image patches, and a random subset of these patches is sampled as inputs to the image encoder. The decoder outputs high-resolution reconstruction results for the down-sampled input image patches. Following MAE, we perform high-resolution reconstruction only on masked patch representations. Therefore, MaCo follows the same training procedure as MAE, with the difference being that the input to MaCo is the down-sampled version of the original images. Let ${v'}_{recon}$ denotes the image decoder's results with input of the down-sampled image, the loss function of MaCo's pretext task is defined as:
\begin{equation}
\centering
\label{Eq.2}
\mathcal{L}_{pret}=||{v'}_{recon}^{msk}-{g}_{recon}^{msk}||^{2}
\end{equation}

\subsubsection*{Report feature extraction}
We adopt BERT \cite{devlin2018bert}, a classical natural language processing model that has achieved good performance across various language understanding tasks, to extract expert knowledge from clinical daily examination reports.

The clinical reports are processed by dividing them into multiple sentences. In this pre-processing step, we also incorporate random sentence selection and shuffling. Next, we use a wordpiece tokenizer to convert the pre-processed reports into a sequence of numerical tokens that can be processed by BERT. The wordpiece tokenizer breaks down each word into sub-word units and maps them to their corresponding numerical representations. This allows BERT to capture the meaning of the text at a more granular level, improving the quality of the sentence representations.

We feed the sequence of numerical tokens into BERT to obtain sentence representations, denoting as $t_{enc}$ with the size of $B \times N^tC$, where $N^{t}$ is the length of text tokens concatenate with the [cls] token. These sentence representations capture the main ideas and themes from the clinical reports and will be used to interact with the extracted image representations, which will be discussed in the next section.

\subsubsection*{Masked contrastive learning with a correlation weighting mechanism}

In this section, our objective is to construct a multi-modal embedding space using sampled image patch representations $v_{enc}$ and report representations $t_{enc}$. The fundamental concept is akin to CLIP \cite{radford2021learning}, wherein a multi-modal embedding space is learned by concurrently training an image encoder and text encoder. Given a batch $B$ of image-report pairs, the goal is to align the image and text in the feature space by maximizing the cosine similarity between the image and text representations of correct image-report pairs while minimizing the cosine similarity of representations for incorrect pairs. 

Let $fc_{i}(\cdot)$ and $fc_{t}(\cdot)$ denote linear mappings in the joint embedding space for image representation and report representation, respectively. Image representations mapping $v = fc_{i}(v_{enc}^{pool})$, and report representations mapping $t = fc_{t}(t_{enc}^{pool})$ is used to calculate the cosine similarity matrix $<v, t>$, where $v_{enc}^{pool}$ with the size of $B \times C$ represents the tokens-dimension pooling result of $v_{enc}$ and $t_{enc}^{pool}$ also with the size of $B \times C$ represents the [cls] token result of $t_{enc}$. With the temperature $\tau$, the InfoNCE loss \cite{sohn2016improved} utilized in a batch is then be described as:

\begin{equation}
\centering
\label{Eq.4}
 \mathcal{L}_{infoNCE}= -\frac{1}{B}\sum_{i}^{B}log(\frac{exp(\langle v_{i},t_{i} \rangle/\tau)}{\sum_{k}^{B}exp(\langle v_{i},t_{k}\rangle/\tau)})
\end{equation}
Here, $\tau$ is optimized during the model training.

However, unlike the common contrastive learning setting with full-resolution full-sampled image inputs, two challenges must be addressed when aligning the multi-modal representations in masked contrastive learning methods: 1) Do the randomly masked images still retain sufficient information that can be correlated with the corresponding reports? 2) If yes, what is the extent of the correlation? Answering these two questions is crucial for establishing meaningful correlations between the image and the text modalities. From the perspective of a clinical expert, the answers to these two questions depend on the quality and relevance of the sampled patches. If the sampled patches can precisely cover the entire lesion area, the two modalities should be highly correlated. Otherwise, the correlation would be low. 

To capture the correlation between paired masked images and reports in a manner that aligns with the expert practice, we propose a correlation weighting mechanism. The details are depicted in Fig. 1 (b). Specifically, we score the sampled images based on a masked position map. These scores are then used to adjust the temperature parameter in contrastive learning and the weights in the contrastive loss function. By doing so, higher weights can be given to highly correlated paired samples during the network learning process, facilitating network optimization.

For the $k^{th}$ ($k=1,..., B$) input instance in a batch, we initiate the process by generating a binary matrix ($p_k \in \mathbb{R}^{\sqrt{N} \times \sqrt{N}}$) based on its patch sampling mask used for masked auto-encoding, assigning a value of 0 to the masked regions and a value of 1 to the sampled regions. This binary matrix is named the masked position map. $p_k$ is then reshaped to a one-dimensional vector $\widehat{p_k} \in \mathbb{R}^{N}$ and a fully connected (FC) layer is learned to generate an importance score for the instance from the reshaped masked position map $\widehat{p_k}$ (Fig. 1 (b)(ii)): $w_k^s=\sum_{i=1}^{N}w_i \cdot \widehat{p_{k,i}}$. Here, $w_i$ is the weight of the FC layer, representing the weight assigned to a specific mask position. Corresponding to all instances in a batch, we obtain the importance score vector $W^s = \{w_k^s\} \in \mathbb{R}^{B}$. Additionally, for the weighting purpose, we employed a softplus activation function to re-scale the range of the importance scores, facilitating more stable training. The final importance scores $W^{c} \in \mathbb{R}^{B}$ are generated as the following:

\begin{equation}
\centering
\label{Eq.5}
 W^{c}= log(1+e^{W^{s}})
\end{equation}

Then, we employ the obtained importance scores $W^{c}$ to weight the image-text sample pairs, ensuring that the model assigns greater attention to pairs with more meaningful sampled content (larger importance scores) during the training process. This weighting process consists of two components, involving the weighting of the cosine similarity matrix $<v,t>$ ($<v,t>$ is also called logits, and in the following, we will use logits to indicate $<v,t>$), and the weighting of loss terms.
The weighting of logits is similar to the use of the reciprocal of the temperature coefficient $\tau$. Generally, a smaller temperature coefficient indicates sharper logits, thereby offering a more rigorous distribution alignment during the training process. In contrast to the temperature coefficient, which has the same value for all sample pairs, our importance scores provide varying weighting values to the digits of different sample pairs in a batch. Particularly, for the $i^{th}$ input image-text sample pair, if the sampled image patches are highly correlated with the corresponding text, a larger importance score (larger $w_i^c$) is obtained, and sharper logits are required. Conversely, when the sampled image patches have a low correlation with the corresponding text, $w_i^c$ is smaller, and relatively uniform distributed logits are learned. In the meantime, we further utilize a detached version of $W^c$ to weight the loss terms, ensuring that samples with higher correlation receive greater attention. 

The proposed masked-contrastive learning loss can thus be expressed as:
\begin{equation}
\centering
\label{Eq.6}
 \mathcal{L}_{contra}= -\frac{1}{B}\sum_{i}^{B}(log(\frac{exp(w_{i}^{c}\cdot \langle v_{i},t_{i} \rangle/ \tau)}{\sum_{k}^{B} exp(w^{c}_{i}\cdot \langle v_{i},t_{k}\rangle/\tau)}) +  w_{i}^{c}log(\frac{exp(\langle v_{i},t_{i} \rangle/\tau)}{\sum_{k}^{B}exp(\langle v_{i},t_{k}\rangle/\tau)}))
\end{equation}

The final loss function to train MaCo combines the pretext-task loss with the masked-contrastive learning loss:
\begin{equation}
\centering
\label{Eq.7}
\mathcal{L}=\lambda\mathcal{L}_{pret} + (1-\lambda)\mathcal{L}_{contra}
\end{equation}
Here, $\lambda$ is a hyperparameter to balance the contributions of the two loss terms.

\subsection*{Implementation details}
We used the same data augmentation methods at different training stages. Specifically, we applied random horizontal flipping, random affine transformations (with degrees set to 20 and scale ranging from 0.8 to 1.2), and normalized the data with a mean of 0.4978 and a standard deviation of 0.2449. All experiments were conducted using the PyTorch framework. The pre-training of MaCo was completed in approximately 3.5 hours using four NVIDIA A100 GPUs. For the sake of convenience and comparability, we utilized the widely-used image encoder ViT-B/16 and employed BERT with a width of 768 as our text encoder. For pre-training, we set the training batch size to 512 and employed the AdamW optimizer, with an initial learning rate of 4.5e-4, weight decay of 0.05, $\beta_{1}$ of 0.9, and $\beta_{2}$ of 0.95. We used a symmetrical design for the contrastive learning loss $\mathcal{L}_{infoNCE}$, following \cite{radford2021learning}. We set the value of $\lambda$ in Eq. \ref{Eq.7} to 0.9. 
The learnable parameter $\tau$ in Eq. \ref{Eq.4} was initialized to 0.03. In fine-tuning tasks, following the practice adopted by the classical methods \cite{9710099, wu2023medklip, zhou2022advancing}, we utilized the pre-trained image encoder as the initialization for the model to be fine-tuned across various applications, including classification, segmentation, and detection.

For the fine-tuning classification experiments on datasets CheXpert, NIH ChestX-ray, and RSNA, we utilized the SGD optimizer, setting its momentum to 0.9 and searching for the optimal learning rate ranging from 8e-3 to 1e-4. For the fine-tuning segmentation experiments on datasets SIIM and COVID Rural, we used the AdamW optimizer, with an initial learning rate of 2e-5, weight decay of 0.05, $\beta_{1}$ of 0.9, and $\beta_{2}$ of 0.999. For the fine-tuning detection experiments on dataset RSNA, we employed VITDet \cite{li2022exploring} as the base detection framework, and we utilized the AdamW optimizer with an initial learning rate of 3e-3, weight decay of 0.1, $\beta_{1}$ of 0.9, and $\beta_{2}$ of 0.999.
    
In both the pre-training and fine-tuning stages of the image classification tasks, we warmed up the network by linearly increasing the learning rate to the set value and then, decreased the learning rate according to the cosine decay schedule. 

\subsection*{Comparative methods}
We began our analysis by comparing MaCo with various pre-training approaches that utilize text as supervision to learn image representations. These approaches include ConVIRT \cite{zhang2022contrastive}, GLoRIA \cite{9710099}, BioViL \cite{boecking2022making}, REFERS \cite{zhou2022generalized}, MGCA \cite{wang2022multi}, MFLAG \cite{liu2023m}, Med-UniC \cite{wan2023med}, M3AE \cite{geng2022multimodal}, MedKLIP \cite{wu2023medklip}, MRM \cite{zhou2022advancing}, LoVT \cite{lovt} and Ark \cite{ma2023foundation}. Specifically, ConVIRT proposes to learn medical visual representations by contrasting paired radiographs and sentences from radiology reports. GLoRIA improves upon ConVIRT by contrasting radiograph patches and words in the reports. BioViL and REFERS incorporate masked language modeling loss into contrastive learning, with REFERS introducing a multi-view fusion attention mechanism to better align the representations of each radiograph and its associated report. M3AE employs mask modeling in both the image and language modalities to investigate the performance of pre-trained models on natural datasets. MedKLIP utilizes a report filter to extract medical entities and employs a more complex modal fusion module to aggregate features. Similar to M3AE, MRM leverages a masking mechanism in both image and text branches, which has achieved the most advanced results in the medical field. To comprehensively evaluate our method, we also introduced some image-based self-supervised learning methods, which include Context Restoration \cite{chen2019self}, Model Genesis \cite{zhou2021models}, TransVW \cite{haghighi2021transferable}, C2L \cite{zhou2020comparing}, and ImageNet \cite{wang2017chestx}.

For the zero-shot tasks, we compared our method with relevant state-of-art approaches, including ConVIRT \cite{zhang2022contrastive}, GLoRIA \cite{9710099}, BioViL \cite{boecking2022making}, CheXzero \cite{tiu2022expert} and MedKLIP \cite{wu2023medklip}. It should be noted that CheXzero and MedKLIP is not capable of handling free-form text, while MRM and M3AE are unable to achieve zero-shot results due to their training strategy. Finally, we demonstrated the weight visualization of our proposed correlation weighting mechanism, where we utilized attention maps to indicate that our approach can weigh the masked image representations in an interpretable and clinically plausible manner.

\subsection*{Datasets}
We pre-train MaCo using radiographs and clinical reports from the MIMIC-CXR V2 dataset \cite{johnson2019mimic}. To assess the transferability of the learned radiograph representations, we perform various X-ray-based downstream tasks using multiple datasets, including NIH ChestX-ray \cite{wang2017chestx}, CheXpert \cite{irvin2019chexpert}, RSNA Pneumonia Detection (RSNA) \cite{shih2019augmenting,wang2017chestx}, SIIM-ACR Pneumothorax \cite{Zawacki2019}, COVID-19 Rural \cite{tang2020deep} dataset and MS-CXR dataset \cite{boecking2022making}, respectively. The following section will introduce the datasets in detail:

MIMIC-CXR v2 is a large dataset comprising 377,110 chest X-rays associated with 227,827 clinical reports sourced from the Beth Israel Deaconess Medical Center between 2011 and 2016. During pre-training, we used all paired data, regardless of whether it was frontal or lateral.

CheXpert releases a multi-label dataset for chest X-ray classification. To evaluate the performance of our model, we followed the official guidelines outlined in \cite{irvin2019chexpert} and reported results for five selected pathologies. As the official CheXpert test set is not publicly available, we adopted a similar approach as described in \cite{zhang2022contrastive} and used the official validation set as our test set. Additionally, following \cite{zhou2022advancing}, we sampled 5,000 images from the official training set to construct our validation set. The resulting training/validation/test split consists of 218,414/5,000/234 images, respectively, representing the entire dataset.

NIH ChestX-ray (NIH) contains 112,120 frontal-view chest radiograph images and focuses on a multi-label classification problem involving 14 different chest pathologies. The dataset is split into training, validation, and test sets, with each set comprising 70$\%$, 10$\%$, and 20$\%$ of the total dataset, respectively.

COVID-19 Rural (COVID Rural) is a small-scale collection comprising over 200 chest X-ray images with COVID-19 disease segmentation masks. We utilize this dataset to evaluate our segmentation performance. The dataset is randomly split into training, validation, and test sets, with a ratio of 60\%, 20\% and 20\%.

SIIM-ACR Pneumothorax (SIIM) is curated to facilitate the development of segmentation models for identifying pneumothorax disease in chest radiographs. The dataset includes more than 120,000 frontal-view chest X-rays, each accompanied by precise manual segmentation of pneumothorax regions. We leverage this dataset for both fine-tuning segmentation and zero-shot classification tasks. In constructing the fine-tuning dataset, our methodology aligns with established practices outlined in \cite{9710099}. Specifically, we partition the dataset into sets for training, validation, and testing, allocating 70$\%$, 15$\%$, and 15$\%$ of the total dataset, respectively.

RSNA Pneumonia Detection (RSNA) is derived from the 2018 RSNA Pneumonia Challenge, comprising a total of 6,012 slices with bounding box annotations. We use this dataset in fine-tuning classification and detection task. For the task of classification, we adhere to the official data split strategy, partitioning the dataset into a training set of 25,184 images, a validation set of 1,500 images, and a test set of 3,000 images. For the task of detection, in alignment with the approach adopted in LoVT \cite{lovt}, the dataset is partitioned into a training set consisting of 3,584 images, a validation set comprising 1,210 images, and a test set with 1,218 images.

MS-CXR provides annotations in the form of bounding boxes and sentence pairs that describe clinical findings observed in chest X-ray images. Each sentence describes a single pathology present in the image, and there could be multiple manually annotated bounding boxes associated with the description of a single radiological finding. The annotations were collected on a subset of MIMIC-CXR images, which contain labels across eight different pathologies. In total, 1,162 annotations of 881 cases were collected, and we utilized the entire dataset to measure the overlap between labeled bounding boxes and the results of vision-language association after pre-training.

\section*{Data availability}
All the data used in this paper are from open-source datasets, including:
MIMIC-CXR v2 (https://physionet.org/content/mimic-cxr-jpg/2.0.0/), 
CheXpert (https://stanfordmlgroup.github.io/competitions/chexpert/), 
NIH (https://www.kaggle.com/datasets/nih-chest-xrays/data), 
COVID Rural (https://github.com/ieee8023/covid-chestxray-dataset), 
SIIM (https://www.kaggle.com/c/siim-acr-pneumothorax-segmentation), 
RSNA (https://www.kaggle.com/c/rsna-pneumonia-detection-challenge/data), 
and MS-CXR (https://physionet.org/content/ms-cxr/0.1/).

\section*{Code Availability}
Our code are available at https://github.com/SZUHvern/MaCo.

\bibliography{sn-bibliography}


\begin{thebibliography}{50}
\ifx \bisbn   \undefined \def \bisbn  #1{ISBN #1}\fi
\ifx \binits  \undefined \def \binits#1{#1}\fi
\ifx \bauthor  \undefined \def \bauthor#1{#1}\fi
\ifx \batitle  \undefined \def \batitle#1{#1}\fi
\ifx \bjtitle  \undefined \def \bjtitle#1{#1}\fi
\ifx \bvolume  \undefined \def \bvolume#1{\textbf{#1}}\fi
\ifx \byear  \undefined \def \byear#1{#1}\fi
\ifx \bissue  \undefined \def \bissue#1{#1}\fi
\ifx \bfpage  \undefined \def \bfpage#1{#1}\fi
\ifx \blpage  \undefined \def \blpage #1{#1}\fi
\ifx \burl  \undefined \def \burl#1{\textsf{#1}}\fi
\ifx \doiurl  \undefined \def \doiurl#1{\url{https://doi.org/#1}}\fi
\ifx \betal  \undefined \def \betal{\textit{et al.}}\fi
\ifx \binstitute  \undefined \def \binstitute#1{#1}\fi
\ifx \binstitutionaled  \undefined \def \binstitutionaled#1{#1}\fi
\ifx \bctitle  \undefined \def \bctitle#1{#1}\fi
\ifx \beditor  \undefined \def \beditor#1{#1}\fi
\ifx \bpublisher  \undefined \def \bpublisher#1{#1}\fi
\ifx \bbtitle  \undefined \def \bbtitle#1{#1}\fi
\ifx \bedition  \undefined \def \bedition#1{#1}\fi
\ifx \bseriesno  \undefined \def \bseriesno#1{#1}\fi
\ifx \blocation  \undefined \def \blocation#1{#1}\fi
\ifx \bsertitle  \undefined \def \bsertitle#1{#1}\fi
\ifx \bsnm \undefined \def \bsnm#1{#1}\fi
\ifx \bsuffix \undefined \def \bsuffix#1{#1}\fi
\ifx \bparticle \undefined \def \bparticle#1{#1}\fi
\ifx \barticle \undefined \def \barticle#1{#1}\fi
\bibcommenthead
\ifx \bconfdate \undefined \def \bconfdate #1{#1}\fi
\ifx \botherref \undefined \def \botherref #1{#1}\fi
\ifx \url \undefined \def \url#1{\textsf{#1}}\fi
\ifx \bchapter \undefined \def \bchapter#1{#1}\fi
\ifx \bbook \undefined \def \bbook#1{#1}\fi
\ifx \bcomment \undefined \def \bcomment#1{#1}\fi
\ifx \oauthor \undefined \def \oauthor#1{#1}\fi
\ifx \citeauthoryear \undefined \def \citeauthoryear#1{#1}\fi
\ifx \endbibitem  \undefined \def \endbibitem {}\fi
\ifx \bconflocation  \undefined \def \bconflocation#1{#1}\fi
\ifx \arxivurl  \undefined \def \arxivurl#1{\textsf{#1}}\fi
\csname PreBibitemsHook\endcsname

\bibitem[\protect\citeauthoryear{Rajpurkar and Lungren}{2023}]{rajpurkar2023current}
\begin{barticle}
\bauthor{\bsnm{Rajpurkar}, \binits{P.}},
\bauthor{\bsnm{Lungren}, \binits{M.P.}}:
\batitle{{T}he {C}urrent and {F}uture {S}tate of {AI} {I}nterpretation of {M}edical {I}mages}.
\bjtitle{New England Journal of Medicine}
\bvolume{388}(\bissue{21}),
\bfpage{1981}--\blpage{1990}
(\byear{2023})
\end{barticle}
\endbibitem

\bibitem[\protect\citeauthoryear{Chang et~al.}{2023}]{chang2023mining}
\begin{barticle}
\bauthor{\bsnm{Chang}, \binits{Q.}},
\bauthor{\bsnm{Yan}, \binits{Z.}},
\bauthor{\bsnm{Zhou}, \binits{M.}},
\bauthor{\bsnm{Qu}, \binits{H.}},
\bauthor{\bsnm{He}, \binits{X.}},
\bauthor{\bsnm{Zhang}, \binits{H.}},
\bauthor{\bsnm{Baskaran}, \binits{L.}},
\bauthor{\bsnm{Al’Aref}, \binits{S.}},
\bauthor{\bsnm{Li}, \binits{H.}},
\bauthor{\bsnm{Zhang}, \binits{S.}}, \betal:
\batitle{Mining multi-center heterogeneous medical data with distributed synthetic learning}.
\bjtitle{Nature Communications}
\bvolume{14}(\bissue{1}),
\bfpage{5510}
(\byear{2023})
\end{barticle}
\endbibitem

\bibitem[\protect\citeauthoryear{Liu et~al.}{2024}]{liu2024swin}
\begin{botherref}
\oauthor{\bsnm{Liu}, \binits{J.}},
\oauthor{\bsnm{Yang}, \binits{H.}},
\oauthor{\bsnm{Zhou}, \binits{H.-Y.}},
\oauthor{\bsnm{Xi}, \binits{Y.}},
\oauthor{\bsnm{Yu}, \binits{L.}},
\oauthor{\bsnm{Yu}, \binits{Y.}},
\oauthor{\bsnm{Liang}, \binits{Y.}},
\oauthor{\bsnm{Shi}, \binits{G.}},
\oauthor{\bsnm{Zhang}, \binits{S.}},
\oauthor{\bsnm{Zheng}, \binits{H.}}, et al.:
Swin-umamba: Mamba-based unet with imagenet-based pretraining.
arXiv preprint arXiv:2402.03302
(2024)
\end{botherref}
\endbibitem

\bibitem[\protect\citeauthoryear{Acosta et~al.}{2022}]{acosta2022multimodal}
\begin{barticle}
\bauthor{\bsnm{Acosta}, \binits{J.N.}},
\bauthor{\bsnm{Falcone}, \binits{G.J.}},
\bauthor{\bsnm{Rajpurkar}, \binits{P.}},
\bauthor{\bsnm{Topol}, \binits{E.J.}}:
\batitle{Multimodal biomedical {AI}}.
\bjtitle{Nature Medicine}
\bvolume{28}(\bissue{9}),
\bfpage{1773}--\blpage{1784}
(\byear{2022})
\end{barticle}
\endbibitem

\bibitem[\protect\citeauthoryear{Moor et~al.}{2023}]{moor2023foundation}
\begin{barticle}
\bauthor{\bsnm{Moor}, \binits{M.}},
\bauthor{\bsnm{Banerjee}, \binits{O.}},
\bauthor{\bsnm{Abad}, \binits{Z.S.H.}},
\bauthor{\bsnm{Krumholz}, \binits{H.M.}},
\bauthor{\bsnm{Leskovec}, \binits{J.}},
\bauthor{\bsnm{Topol}, \binits{E.J.}},
\bauthor{\bsnm{Rajpurkar}, \binits{P.}}:
\batitle{Foundation models for generalist medical artificial intelligence}.
\bjtitle{Nature}
\bvolume{616}(\bissue{7956}),
\bfpage{259}--\blpage{265}
(\byear{2023})
\end{barticle}
\endbibitem

\bibitem[\protect\citeauthoryear{Wu et~al.}{2023}]{wu2023medklip}
\begin{bchapter}
\bauthor{\bsnm{Wu}, \binits{C.}},
\bauthor{\bsnm{Zhang}, \binits{X.}},
\bauthor{\bsnm{Zhang}, \binits{Y.}},
\bauthor{\bsnm{Wang}, \binits{Y.}},
\bauthor{\bsnm{Xie}, \binits{W.}}:
\bctitle{Medklip: Medical knowledge enhanced language-image pre-training for x-ray diagnosis}.
In: \bbtitle{Proceedings of the IEEE/CVF International Conference on Computer Vision},
pp. \bfpage{21372}--\blpage{21383}
(\byear{2023})
\end{bchapter}
\endbibitem

\bibitem[\protect\citeauthoryear{Tiu et~al.}{2022}]{tiu2022expert}
\begin{barticle}
\bauthor{\bsnm{Tiu}, \binits{E.}},
\bauthor{\bsnm{Talius}, \binits{E.}},
\bauthor{\bsnm{Patel}, \binits{P.}},
\bauthor{\bsnm{Langlotz}, \binits{C.P.}},
\bauthor{\bsnm{Ng}, \binits{A.Y.}},
\bauthor{\bsnm{Rajpurkar}, \binits{P.}}:
\batitle{Expert-level detection of pathologies from unannotated chest x-ray images via self-supervised learning}.
\bjtitle{Nature Biomedical Engineering}
\bvolume{6}(\bissue{12}),
\bfpage{1399}--\blpage{1406}
(\byear{2022})
\end{barticle}
\endbibitem

\bibitem[\protect\citeauthoryear{Liu et~al.}{2024}]{isbiliu}
\begin{bchapter}
\bauthor{\bsnm{Liu}, \binits{J.}},
\bauthor{\bsnm{Zhou}, \binits{H.-Y.}},
\bauthor{\bsnm{Li}, \binits{C.}},
\bauthor{\bsnm{Huang}, \binits{W.}},
\bauthor{\bsnm{Yang}, \binits{H.}},
\bauthor{\bsnm{Liang}, \binits{Y.}},
\bauthor{\bsnm{Wang}, \binits{S.}}:
\bctitle{Mlip: Medical language-image pre-training with masked local representation learning}.
In: \bbtitle{2024 IEEE 21th International Symposium on Biomedical Imaging (ISBI)}
(\byear{2024})
\end{bchapter}
\endbibitem

\bibitem[\protect\citeauthoryear{Zhou et~al.}{2023}]{zhou2023foundation}
\begin{botherref}
\oauthor{\bsnm{Zhou}, \binits{Y.}},
\oauthor{\bsnm{Chia}, \binits{M.A.}},
\oauthor{\bsnm{Wagner}, \binits{S.K.}},
\oauthor{\bsnm{Ayhan}, \binits{M.S.}},
\oauthor{\bsnm{Williamson}, \binits{D.J.}},
\oauthor{\bsnm{Struyven}, \binits{R.R.}},
\oauthor{\bsnm{Liu}, \binits{T.}},
\oauthor{\bsnm{Xu}, \binits{M.}},
\oauthor{\bsnm{Lozano}, \binits{M.G.}},
\oauthor{\bsnm{Woodward-Court}, \binits{P.}}, et al.:
A foundation model for generalizable disease detection from retinal images.
Nature,
1--8
(2023)
\end{botherref}
\endbibitem

\bibitem[\protect\citeauthoryear{Yang et~al.}{2024}]{isbiyang}
\begin{bchapter}
\bauthor{\bsnm{Yang}, \binits{H.}},
\bauthor{\bsnm{Zhou}, \binits{H.-Y.}},
\bauthor{\bsnm{Li}, \binits{C.}},
\bauthor{\bsnm{Huang}, \binits{W.}},
\bauthor{\bsnm{Liu}, \binits{J.}},
\bauthor{\bsnm{Liang}, \binits{Y.}},
\bauthor{\bsnm{Wang}, \binits{S.}}:
\bctitle{Multimodal self-supervised learning for lesion localization}.
In: \bbtitle{2024 IEEE 21th International Symposium on Biomedical Imaging (ISBI)}
(\byear{2024})
\end{bchapter}
\endbibitem

\bibitem[\protect\citeauthoryear{Zhou et~al.}{2023}]{zhou2023unified}
\begin{barticle}
\bauthor{\bsnm{Zhou}, \binits{H.-Y.}}, \betal:
\batitle{A {U}nified {V}isual {I}nformation {P}reservation {F}ramework for {S}elf-supervised {P}re-training in {M}edical {I}mage {A}nalysis}.
\bjtitle{IEEE Transactions on Pattern Analysis and Machine Intelligence}
\bvolume{45}(\bissue{7}),
\bfpage{8020}--\blpage{8035}
(\byear{2023})
\doiurl{10.1109/TPAMI.2023.3234002}
\end{barticle}
\endbibitem

\bibitem[\protect\citeauthoryear{Zhou et~al.}{2022}]{zhou2022generalized}
\begin{barticle}
\bauthor{\bsnm{Zhou}, \binits{H.-Y.}},
\bauthor{\bsnm{Chen}, \binits{X.}},
\bauthor{\bsnm{Zhang}, \binits{Y.}},
\bauthor{\bsnm{Luo}, \binits{R.}},
\bauthor{\bsnm{Wang}, \binits{L.}},
\bauthor{\bsnm{Yu}, \binits{Y.}}:
\batitle{Generalized radiograph representation learning via cross-supervision between images and free-text radiology reports}.
\bjtitle{Nature Machine Intelligence}
\bvolume{4}(\bissue{1}),
\bfpage{32}--\blpage{40}
(\byear{2022})
\end{barticle}
\endbibitem

\bibitem[\protect\citeauthoryear{Huang et~al.}{2024}]{isbihuang}
\begin{bchapter}
\bauthor{\bsnm{Huang}, \binits{W.}},
\bauthor{\bsnm{Li}, \binits{C.}},
\bauthor{\bsnm{Zhou}, \binits{H.-Y.}},
\bauthor{\bsnm{Liu}, \binits{J.}},
\bauthor{\bsnm{Yang}, \binits{H.}},
\bauthor{\bsnm{Liang}, \binits{Y.}},
\bauthor{\bsnm{Shi}, \binits{G.}},
\bauthor{\bsnm{Zheng}, \binits{H.}},
\bauthor{\bsnm{Wang}, \binits{S.}}:
\bctitle{Enhancing representation in medical vision-language foundation models via multi-scale information extraction techniques}.
In: \bbtitle{2024 IEEE 21th International Symposium on Biomedical Imaging (ISBI)}
(\byear{2024})
\end{bchapter}
\endbibitem

\bibitem[\protect\citeauthoryear{He et~al.}{2022}]{9879206}
\begin{bchapter}
\bauthor{\bsnm{He}, \binits{K.}},
\bauthor{\bsnm{Chen}, \binits{X.}},
\bauthor{\bsnm{Xie}, \binits{S.}},
\bauthor{\bsnm{Li}, \binits{Y.}},
\bauthor{\bsnm{Dollár}, \binits{P.}},
\bauthor{\bsnm{Girshick}, \binits{R.}}:
\bctitle{Masked {A}utoencoders {A}re {S}calable {V}ision {L}earners}.
In: \bbtitle{2022 IEEE/CVF Conference on Computer Vision and Pattern Recognition (CVPR)},
pp. \bfpage{15979}--\blpage{15988}
(\byear{2022}).
\doiurl{10.1109/CVPR52688.2022.01553}
\end{bchapter}
\endbibitem

\bibitem[\protect\citeauthoryear{Sutton et~al.}{2020}]{sutton2020overview}
\begin{barticle}
\bauthor{\bsnm{Sutton}, \binits{R.T.}},
\bauthor{\bsnm{Pincock}, \binits{D.}},
\bauthor{\bsnm{Baumgart}, \binits{D.C.}},
\bauthor{\bsnm{Sadowski}, \binits{D.C.}},
\bauthor{\bsnm{Fedorak}, \binits{R.N.}},
\bauthor{\bsnm{Kroeker}, \binits{K.I.}}:
\batitle{An overview of clinical decision support systems: benefits, risks, and strategies for success}.
\bjtitle{NPJ digital medicine}
\bvolume{3}(\bissue{1}),
\bfpage{17}
(\byear{2020})
\end{barticle}
\endbibitem

\bibitem[\protect\citeauthoryear{Zhang et~al.}{2023}]{zhang2023knowledge}
\begin{barticle}
\bauthor{\bsnm{Zhang}, \binits{X.}},
\bauthor{\bsnm{Wu}, \binits{C.}},
\bauthor{\bsnm{Zhang}, \binits{Y.}},
\bauthor{\bsnm{Xie}, \binits{W.}},
\bauthor{\bsnm{Wang}, \binits{Y.}}:
\batitle{Knowledge-enhanced visual-language pre-training on chest radiology images}.
\bjtitle{Nature Communications}
\bvolume{14}(\bissue{1}),
\bfpage{4542}
(\byear{2023})
\end{barticle}
\endbibitem

\bibitem[\protect\citeauthoryear{Zhou et~al.}{2023}]{zhou2023transformer}
\begin{botherref}
\oauthor{\bsnm{Zhou}, \binits{H.-Y.}},
\oauthor{\bsnm{Yu}, \binits{Y.}},
\oauthor{\bsnm{Wang}, \binits{C.}},
\oauthor{\bsnm{Zhang}, \binits{S.}},
\oauthor{\bsnm{Gao}, \binits{Y.}},
\oauthor{\bsnm{Pan}, \binits{J.}},
\oauthor{\bsnm{Shao}, \binits{J.}},
\oauthor{\bsnm{Lu}, \binits{G.}},
\oauthor{\bsnm{Zhang}, \binits{K.}},
\oauthor{\bsnm{Li}, \binits{W.}}:
A transformer-based representation-learning model with unified processing of multimodal input for clinical diagnostics.
Nature Biomedical Engineering,
1--13
(2023)
\end{botherref}
\endbibitem

\bibitem[\protect\citeauthoryear{Huang et~al.}{2023}]{huang2023visual}
\begin{botherref}
\oauthor{\bsnm{Huang}, \binits{Z.}},
\oauthor{\bsnm{Bianchi}, \binits{F.}},
\oauthor{\bsnm{Yuksekgonul}, \binits{M.}},
\oauthor{\bsnm{Montine}, \binits{T.J.}},
\oauthor{\bsnm{Zou}, \binits{J.}}:
A visual--language foundation model for pathology image analysis using medical twitter.
Nature Medicine,
1--10
(2023)
\end{botherref}
\endbibitem

\bibitem[\protect\citeauthoryear{Zhou et~al.}{2023}]{zhou2022advancing}
\begin{bchapter}
\bauthor{\bsnm{Zhou}, \binits{H.-Y.}},
\bauthor{\bsnm{Lian}, \binits{C.}},
\bauthor{\bsnm{Wang}, \binits{L.}},
\bauthor{\bsnm{Yu}, \binits{Y.}}:
\bctitle{Advancing {R}adiograph {R}epresentation {L}earning with {M}asked {R}ecord {M}odeling}.
In: \bbtitle{The Eleventh International Conference on Learning Representations}
(\byear{2023})
\end{bchapter}
\endbibitem

\bibitem[\protect\citeauthoryear{Chen et~al.}{2022}]{chen2022multi}
\begin{bchapter}
\bauthor{\bsnm{Chen}, \binits{Z.}},
\bauthor{\bsnm{Du}, \binits{Y.}},
\bauthor{\bsnm{Hu}, \binits{J.}},
\bauthor{\bsnm{Liu}, \binits{Y.}},
\bauthor{\bsnm{Li}, \binits{G.}},
\bauthor{\bsnm{Wan}, \binits{X.}},
\bauthor{\bsnm{Chang}, \binits{T.-H.}}:
\bctitle{Multi-modal masked autoencoders for medical vision-and-language pre-training}.
In: \bbtitle{International Conference on Medical Image Computing and Computer-Assisted Intervention},
pp. \bfpage{679}--\blpage{689}
(\byear{2022}).
\bcomment{Springer}
\end{bchapter}
\endbibitem

\bibitem[\protect\citeauthoryear{Radford et~al.}{2021}]{radford2021learning}
\begin{bchapter}
\bauthor{\bsnm{Radford}, \binits{A.}},
\bauthor{\bsnm{Kim}, \binits{J.W.}},
\bauthor{\bsnm{Hallacy}, \binits{C.}},
\bauthor{\bsnm{Ramesh}, \binits{A.}},
\bauthor{\bsnm{Goh}, \binits{G.}},
\bauthor{\bsnm{Agarwal}, \binits{S.}},
\bauthor{\bsnm{Sastry}, \binits{G.}},
\bauthor{\bsnm{Askell}, \binits{A.}},
\bauthor{\bsnm{Mishkin}, \binits{P.}},
\bauthor{\bsnm{Clark}, \binits{J.}}, \betal:
\bctitle{Learning transferable visual models from natural language supervision}.
In: \bbtitle{International Conference on Machine Learning},
pp. \bfpage{8748}--\blpage{8763}
(\byear{2021}).
\bcomment{PMLR}
\end{bchapter}
\endbibitem

\bibitem[\protect\citeauthoryear{Huang et~al.}{2021}]{9710099}
\begin{bchapter}
\bauthor{\bsnm{Huang}, \binits{S.-C.}},
\bauthor{\bsnm{Shen}, \binits{L.}},
\bauthor{\bsnm{Lungren}, \binits{M.P.}},
\bauthor{\bsnm{Yeung}, \binits{S.}}:
\bctitle{{GL}o{RIA}: {A} {M}ultimodal {G}lobal-{L}ocal {R}epresentation {L}earning {F}ramework for {L}abel-efficient {M}edical {I}mage {R}ecognition}.
In: \bbtitle{2021 IEEE/CVF International Conference on Computer Vision (ICCV)},
pp. \bfpage{3922}--\blpage{3931}
(\byear{2021}).
\doiurl{10.1109/ICCV48922.2021.00391}
\end{bchapter}
\endbibitem

\bibitem[\protect\citeauthoryear{Boecking et~al.}{2022}]{boecking2022making}
\begin{bchapter}
\bauthor{\bsnm{Boecking}, \binits{B.}},
\bauthor{\bsnm{Usuyama}, \binits{N.}},
\bauthor{\bsnm{Bannur}, \binits{S.}},
\bauthor{\bsnm{Castro}, \binits{D.C.}},
\bauthor{\bsnm{Schwaighofer}, \binits{A.}},
\bauthor{\bsnm{Hyland}, \binits{S.}},
\bauthor{\bsnm{Wetscherek}, \binits{M.}},
\bauthor{\bsnm{Naumann}, \binits{T.}},
\bauthor{\bsnm{Nori}, \binits{A.}},
\bauthor{\bsnm{Alvarez-Valle}, \binits{J.}}, \betal:
\bctitle{Making the most of text semantics to improve biomedical vision--language processing}.
In: \bbtitle{European Conference on Computer Vision},
pp. \bfpage{1}--\blpage{21}
(\byear{2022}).
\bcomment{Springer}
\end{bchapter}
\endbibitem

\bibitem[\protect\citeauthoryear{M{\"u}ller et~al.}{2022}]{lovt}
\begin{bchapter}
\bauthor{\bsnm{M{\"u}ller}, \binits{P.}},
\bauthor{\bsnm{Kaissis}, \binits{G.}},
\bauthor{\bsnm{Zou}, \binits{C.}},
\bauthor{\bsnm{Rueckert}, \binits{D.}}:
\bctitle{Joint learning of localized representations from medical images and reports}.
In: \bbtitle{European Conference on Computer Vision},
pp. \bfpage{685}--\blpage{701}
(\byear{2022}).
\bcomment{Springer}
\end{bchapter}
\endbibitem

\bibitem[\protect\citeauthoryear{Zhou et~al.}{2021}]{zhou2021models}
\begin{barticle}
\bauthor{\bsnm{Zhou}, \binits{Z.}},
\bauthor{\bsnm{Sodha}, \binits{V.}},
\bauthor{\bsnm{Pang}, \binits{J.}},
\bauthor{\bsnm{Gotway}, \binits{M.B.}},
\bauthor{\bsnm{Liang}, \binits{J.}}:
\batitle{Models genesis}.
\bjtitle{Medical image analysis}
\bvolume{67},
\bfpage{101840}
(\byear{2021})
\end{barticle}
\endbibitem

\bibitem[\protect\citeauthoryear{Zhou et~al.}{2020}]{zhou2020comparing}
\begin{bchapter}
\bauthor{\bsnm{Zhou}, \binits{H.-Y.}},
\bauthor{\bsnm{Yu}, \binits{S.}},
\bauthor{\bsnm{Bian}, \binits{C.}},
\bauthor{\bsnm{Hu}, \binits{Y.}},
\bauthor{\bsnm{Ma}, \binits{K.}},
\bauthor{\bsnm{Zheng}, \binits{Y.}}:
\bctitle{Comparing to learn: Surpassing imagenet pretraining on radiographs by comparing image representations}.
In: \bbtitle{Medical Image Computing and Computer Assisted Intervention--MICCAI 2020: 23rd International Conference, Lima, Peru, October 4--8, 2020, Proceedings, Part I 23},
pp. \bfpage{398}--\blpage{407}
(\byear{2020}).
\bcomment{Springer}
\end{bchapter}
\endbibitem

\bibitem[\protect\citeauthoryear{Chen et~al.}{2019}]{chen2019self}
\begin{barticle}
\bauthor{\bsnm{Chen}, \binits{L.}},
\bauthor{\bsnm{Bentley}, \binits{P.}},
\bauthor{\bsnm{Mori}, \binits{K.}},
\bauthor{\bsnm{Misawa}, \binits{K.}},
\bauthor{\bsnm{Fujiwara}, \binits{M.}},
\bauthor{\bsnm{Rueckert}, \binits{D.}}:
\batitle{Self-supervised learning for medical image analysis using image context restoration}.
\bjtitle{Medical image analysis}
\bvolume{58},
\bfpage{101539}
(\byear{2019})
\end{barticle}
\endbibitem

\bibitem[\protect\citeauthoryear{Haghighi et~al.}{2021}]{haghighi2021transferable}
\begin{barticle}
\bauthor{\bsnm{Haghighi}, \binits{F.}},
\bauthor{\bsnm{Taher}, \binits{M.R.H.}},
\bauthor{\bsnm{Zhou}, \binits{Z.}},
\bauthor{\bsnm{Gotway}, \binits{M.B.}},
\bauthor{\bsnm{Liang}, \binits{J.}}:
\batitle{Transferable visual words: Exploiting the semantics of anatomical patterns for self-supervised learning}.
\bjtitle{IEEE transactions on medical imaging}
\bvolume{40}(\bissue{10}),
\bfpage{2857}--\blpage{2868}
(\byear{2021})
\end{barticle}
\endbibitem

\bibitem[\protect\citeauthoryear{Li et~al.}{2022}]{li2022exploring}
\begin{bchapter}
\bauthor{\bsnm{Li}, \binits{Y.}},
\bauthor{\bsnm{Mao}, \binits{H.}},
\bauthor{\bsnm{Girshick}, \binits{R.}},
\bauthor{\bsnm{He}, \binits{K.}}:
\bctitle{Exploring plain vision transformer backbones for object detection}.
In: \bbtitle{European Conference on Computer Vision},
pp. \bfpage{280}--\blpage{296}
(\byear{2022}).
\bcomment{Springer}
\end{bchapter}
\endbibitem

\bibitem[\protect\citeauthoryear{He et~al.}{2016}]{he2016deep}
\begin{bchapter}
\bauthor{\bsnm{He}, \binits{K.}},
\bauthor{\bsnm{Zhang}, \binits{X.}},
\bauthor{\bsnm{Ren}, \binits{S.}},
\bauthor{\bsnm{Sun}, \binits{J.}}:
\bctitle{Deep residual learning for image recognition}.
In: \bbtitle{Proceedings of the IEEE Conference on Computer Vision and Pattern Recognition},
pp. \bfpage{770}--\blpage{778}
(\byear{2016})
\end{bchapter}
\endbibitem

\bibitem[\protect\citeauthoryear{Misra and Maaten}{2020}]{misra2020self}
\begin{bchapter}
\bauthor{\bsnm{Misra}, \binits{I.}},
\bauthor{\bsnm{Maaten}, \binits{L.v.d.}}:
\bctitle{Self-supervised learning of pretext-invariant representations}.
In: \bbtitle{Proceedings of the IEEE/CVF Conference on Computer Vision and Pattern Recognition},
pp. \bfpage{6707}--\blpage{6717}
(\byear{2020})
\end{bchapter}
\endbibitem

\bibitem[\protect\citeauthoryear{Albelwi}{2022}]{albelwi2022survey}
\begin{barticle}
\bauthor{\bsnm{Albelwi}, \binits{S.}}:
\batitle{Survey on self-supervised learning: auxiliary pretext tasks and contrastive learning methods in imaging}.
\bjtitle{Entropy}
\bvolume{24}(\bissue{4}),
\bfpage{551}
(\byear{2022})
\end{barticle}
\endbibitem

\bibitem[\protect\citeauthoryear{Geng et~al.}{2022}]{geng2022multimodal}
\begin{bchapter}
\bauthor{\bsnm{Geng}, \binits{X.}},
\bauthor{\bsnm{Liu}, \binits{H.}},
\bauthor{\bsnm{Lee}, \binits{L.}},
\bauthor{\bsnm{Schuurmans}, \binits{D.}},
\bauthor{\bsnm{Levine}, \binits{S.}},
\bauthor{\bsnm{Abbeel}, \binits{P.}}:
\bctitle{Multimodal {M}asked {A}utoencoders {L}earn {T}ransferable {R}epresentations}.
In: \bbtitle{First Workshop on Pre-training: Perspectives, Pitfalls, and Paths Forward at ICML 2022}
(\byear{2022})
\end{bchapter}
\endbibitem

\bibitem[\protect\citeauthoryear{Jaiswal et~al.}{2020}]{jaiswal2020survey}
\begin{barticle}
\bauthor{\bsnm{Jaiswal}, \binits{A.}},
\bauthor{\bsnm{Babu}, \binits{A.R.}},
\bauthor{\bsnm{Zadeh}, \binits{M.Z.}},
\bauthor{\bsnm{Banerjee}, \binits{D.}},
\bauthor{\bsnm{Makedon}, \binits{F.}}:
\batitle{A survey on contrastive self-supervised learning}.
\bjtitle{Technologies}
\bvolume{9}(\bissue{1}),
\bfpage{2}
(\byear{2020})
\end{barticle}
\endbibitem

\bibitem[\protect\citeauthoryear{Zhang et~al.}{2022}]{zhang2022contrastive}
\begin{bchapter}
\bauthor{\bsnm{Zhang}, \binits{Y.}},
\bauthor{\bsnm{Jiang}, \binits{H.}},
\bauthor{\bsnm{Miura}, \binits{Y.}},
\bauthor{\bsnm{Manning}, \binits{C.D.}},
\bauthor{\bsnm{Langlotz}, \binits{C.P.}}:
\bctitle{Contrastive learning of medical visual representations from paired images and text}.
In: \bbtitle{Machine Learning for Healthcare Conference},
pp. \bfpage{2}--\blpage{25}
(\byear{2022}).
\bcomment{PMLR}
\end{bchapter}
\endbibitem

\bibitem[\protect\citeauthoryear{Wang et~al.}{2022}]{wang2022medclip}
\begin{bchapter}
\bauthor{\bsnm{Wang}, \binits{Z.}},
\bauthor{\bsnm{Wu}, \binits{Z.}},
\bauthor{\bsnm{Agarwal}, \binits{D.}},
\bauthor{\bsnm{Sun}, \binits{J.}}:
\bctitle{Med{CLIP}: Contrastive {L}earning from {U}npaired {M}edical {I}mages and {T}ext}.
In: \bbtitle{Proceedings of the 2022 Conference on Empirical Methods in Natural Language Processing},
pp. \bfpage{3876}--\blpage{3887}
(\byear{2022})
\end{bchapter}
\endbibitem

\bibitem[\protect\citeauthoryear{Bannur et~al.}{2023}]{bannur2023learning}
\begin{bchapter}
\bauthor{\bsnm{Bannur}, \binits{S.}},
\bauthor{\bsnm{Hyland}, \binits{S.}},
\bauthor{\bsnm{Liu}, \binits{Q.}},
\bauthor{\bsnm{Perez-Garcia}, \binits{F.}},
\bauthor{\bsnm{Ilse}, \binits{M.}},
\bauthor{\bsnm{Castro}, \binits{D.C.}},
\bauthor{\bsnm{Boecking}, \binits{B.}},
\bauthor{\bsnm{Sharma}, \binits{H.}},
\bauthor{\bsnm{Bouzid}, \binits{K.}},
\bauthor{\bsnm{Thieme}, \binits{A.}}, \betal:
\bctitle{Learning to exploit temporal structure for biomedical vision-language processing}.
In: \bbtitle{Proceedings of the IEEE/CVF Conference on Computer Vision and Pattern Recognition},
pp. \bfpage{15016}--\blpage{15027}
(\byear{2023})
\end{bchapter}
\endbibitem

\bibitem[\protect\citeauthoryear{Li et~al.}{2023}]{li2023unify}
\begin{bchapter}
\bauthor{\bsnm{Li}, \binits{Y.}},
\bauthor{\bsnm{Yang}, \binits{B.}},
\bauthor{\bsnm{Cheng}, \binits{X.}},
\bauthor{\bsnm{Zhu}, \binits{Z.}},
\bauthor{\bsnm{Li}, \binits{H.}},
\bauthor{\bsnm{Zou}, \binits{Y.}}:
\bctitle{Unify, align and refine: Multi-level semantic alignment for radiology report generation}.
In: \bbtitle{Proceedings of the IEEE/CVF International Conference on Computer Vision},
pp. \bfpage{2863}--\blpage{2874}
(\byear{2023})
\end{bchapter}
\endbibitem

\bibitem[\protect\citeauthoryear{Wang et~al.}{2022}]{wang2022multi}
\begin{barticle}
\bauthor{\bsnm{Wang}, \binits{F.}},
\bauthor{\bsnm{Zhou}, \binits{Y.}},
\bauthor{\bsnm{Wang}, \binits{S.}},
\bauthor{\bsnm{Vardhanabhuti}, \binits{V.}},
\bauthor{\bsnm{Yu}, \binits{L.}}:
\batitle{Multi-granularity cross-modal alignment for generalized medical visual representation learning}.
\bjtitle{Advances in Neural Information Processing Systems}
\bvolume{35},
\bfpage{33536}--\blpage{33549}
(\byear{2022})
\end{barticle}
\endbibitem

\bibitem[\protect\citeauthoryear{Kenton and Toutanova}{2019}]{devlin2018bert}
\begin{bchapter}
\bauthor{\bsnm{Kenton}, \binits{J.D.M.-W.C.}},
\bauthor{\bsnm{Toutanova}, \binits{L.K.}}:
\bctitle{{BERT}: Pre-training of deep bidirectional transformers for language understanding}.
In: \bbtitle{Proceedings of naacL-HLT},
vol. \bseriesno{1},
p. \bfpage{2}
(\byear{2019})
\end{bchapter}
\endbibitem

\bibitem[\protect\citeauthoryear{Sohn}{2016}]{sohn2016improved}
\begin{botherref}
\oauthor{\bsnm{Sohn}, \binits{K.}}:
Improved deep metric learning with multi-class n-pair loss objective.
Advances in neural information processing systems
\textbf{29}
(2016)
\end{botherref}
\endbibitem

\bibitem[\protect\citeauthoryear{Liu et~al.}{2023}]{liu2023m}
\begin{bchapter}
\bauthor{\bsnm{Liu}, \binits{C.}},
\bauthor{\bsnm{Cheng}, \binits{S.}},
\bauthor{\bsnm{Chen}, \binits{C.}},
\bauthor{\bsnm{Qiao}, \binits{M.}},
\bauthor{\bsnm{Zhang}, \binits{W.}},
\bauthor{\bsnm{Shah}, \binits{A.}},
\bauthor{\bsnm{Bai}, \binits{W.}},
\bauthor{\bsnm{Arcucci}, \binits{R.}}:
\bctitle{{M-FLAG}: {M}edical {V}ision-{L}anguage {P}re-training with {F}rozen {L}anguage {M}odels and {L}atent {S}pace {G}eometry {O}ptimization}.
In: \bbtitle{International Conference on Medical Image Computing and Computer-Assisted Intervention},
pp. \bfpage{637}--\blpage{647}
(\byear{2023})
\end{bchapter}
\endbibitem

\bibitem[\protect\citeauthoryear{Wan et~al.}{2024}]{wan2023med}
\begin{botherref}
\oauthor{\bsnm{Wan}, \binits{Z.}},
\oauthor{\bsnm{Liu}, \binits{C.}},
\oauthor{\bsnm{Zhang}, \binits{M.}},
\oauthor{\bsnm{Fu}, \binits{J.}},
\oauthor{\bsnm{Wang}, \binits{B.}},
\oauthor{\bsnm{Cheng}, \binits{S.}},
\oauthor{\bsnm{Ma}, \binits{L.}},
\oauthor{\bsnm{Quilodr{\'a}n-Casas}, \binits{C.}},
\oauthor{\bsnm{Arcucci}, \binits{R.}}:
Med-unic: Unifying cross-lingual medical vision-language pre-training by diminishing bias.
Advances in Neural Information Processing Systems
\textbf{36}
(2024)
\end{botherref}
\endbibitem

\bibitem[\protect\citeauthoryear{Ma et~al.}{2023}]{ma2023foundation}
\begin{bchapter}
\bauthor{\bsnm{Ma}, \binits{D.}},
\bauthor{\bsnm{Pang}, \binits{J.}},
\bauthor{\bsnm{Gotway}, \binits{M.B.}},
\bauthor{\bsnm{Liang}, \binits{J.}}:
\bctitle{Foundation {A}rk: Accruing and {R}eusing {K}nowledge for {S}uperior and {R}obust {P}erformance}.
In: \bbtitle{International Conference on Medical Image Computing and Computer-Assisted Intervention},
pp. \bfpage{651}--\blpage{662}
(\byear{2023}).
\bcomment{Springer}
\end{bchapter}
\endbibitem

\bibitem[\protect\citeauthoryear{Wang et~al.}{2017}]{wang2017chestx}
\begin{bchapter}
\bauthor{\bsnm{Wang}, \binits{X.}},
\bauthor{\bsnm{Peng}, \binits{Y.}},
\bauthor{\bsnm{Lu}, \binits{L.}},
\bauthor{\bsnm{Lu}, \binits{Z.}},
\bauthor{\bsnm{Bagheri}, \binits{M.}},
\bauthor{\bsnm{Summers}, \binits{R.M.}}:
\bctitle{Chestx-ray8: Hospital-scale chest x-ray database and benchmarks on weakly-supervised classification and localization of common thorax diseases}.
In: \bbtitle{Proceedings of the IEEE Conference on Computer Vision and Pattern Recognition},
pp. \bfpage{2097}--\blpage{2106}
(\byear{2017})
\end{bchapter}
\endbibitem

\bibitem[\protect\citeauthoryear{Johnson et~al.}{2019}]{johnson2019mimic}
\begin{barticle}
\bauthor{\bsnm{Johnson}, \binits{A.E.}},
\bauthor{\bsnm{Pollard}, \binits{T.J.}},
\bauthor{\bsnm{Berkowitz}, \binits{S.J.}},
\bauthor{\bsnm{Greenbaum}, \binits{N.R.}},
\bauthor{\bsnm{Lungren}, \binits{M.P.}},
\bauthor{\bsnm{Deng}, \binits{C.-y.}},
\bauthor{\bsnm{Mark}, \binits{R.G.}},
\bauthor{\bsnm{Horng}, \binits{S.}}:
\batitle{Mimic-cxr, a de-identified publicly available database of chest radiographs with free-text reports}.
\bjtitle{Scientific data}
\bvolume{6}(\bissue{1}),
\bfpage{317}
(\byear{2019})
\end{barticle}
\endbibitem

\bibitem[\protect\citeauthoryear{Irvin et~al.}{2019}]{irvin2019chexpert}
\begin{bchapter}
\bauthor{\bsnm{Irvin}, \binits{J.}},
\bauthor{\bsnm{Rajpurkar}, \binits{P.}},
\bauthor{\bsnm{Ko}, \binits{M.}},
\bauthor{\bsnm{Yu}, \binits{Y.}},
\bauthor{\bsnm{Ciurea-Ilcus}, \binits{S.}},
\bauthor{\bsnm{Chute}, \binits{C.}},
\bauthor{\bsnm{Marklund}, \binits{H.}},
\bauthor{\bsnm{Haghgoo}, \binits{B.}},
\bauthor{\bsnm{Ball}, \binits{R.}},
\bauthor{\bsnm{Shpanskaya}, \binits{K.}}, \betal:
\bctitle{Chexpert: A large chest radiograph dataset with uncertainty labels and expert comparison}.
In: \bbtitle{Proceedings of the AAAI Conference on Artificial Intelligence},
vol. \bseriesno{33},
pp. \bfpage{590}--\blpage{597}
(\byear{2019})
\end{bchapter}
\endbibitem

\bibitem[\protect\citeauthoryear{Shih et~al.}{2019}]{shih2019augmenting}
\begin{barticle}
\bauthor{\bsnm{Shih}, \binits{G.}},
\bauthor{\bsnm{Wu}, \binits{C.C.}},
\bauthor{\bsnm{Halabi}, \binits{S.S.}},
\bauthor{\bsnm{Kohli}, \binits{M.D.}},
\bauthor{\bsnm{Prevedello}, \binits{L.M.}},
\bauthor{\bsnm{Cook}, \binits{T.S.}},
\bauthor{\bsnm{Sharma}, \binits{A.}},
\bauthor{\bsnm{Amorosa}, \binits{J.K.}},
\bauthor{\bsnm{Arteaga}, \binits{V.}},
\bauthor{\bsnm{Galperin-Aizenberg}, \binits{M.}}, \betal:
\batitle{Augmenting the national institutes of health chest radiograph dataset with expert annotations of possible pneumonia}.
\bjtitle{Radiology: Artificial Intelligence}
\bvolume{1}(\bissue{1}),
\bfpage{180041}
(\byear{2019})
\end{barticle}
\endbibitem

\bibitem[\protect\citeauthoryear{Zawacki et~al.}{2019}]{Zawacki2019}
\begin{botherref}
\oauthor{\bsnm{Zawacki}, \binits{A.}},
\oauthor{\bsnm{Wu}, \binits{C.}},
\oauthor{\bsnm{Shih}, \binits{G.}},
\oauthor{\bsnm{Elliott}, \binits{J.}},
\oauthor{\bsnm{Fomitchev}, \binits{M.}},
\oauthor{\bsnm{Hussain}, \binits{M.}},
\oauthor{\bsnm{ParasLakhani}},
\oauthor{\bsnm{Culliton}, \binits{P.}},
\oauthor{\bsnm{Bao}, \binits{S.}}:
SIIM-ACR Pneumothorax Segmentation.
https://kaggle.com/competitions/siim-acr-pneumothorax-segmentation
(2019)
\end{botherref}
\endbibitem

\bibitem[\protect\citeauthoryear{Tang et~al.}{2020}]{tang2020deep}
\begin{botherref}
\oauthor{\bsnm{Tang}, \binits{H.}},
\oauthor{\bsnm{Sun}, \binits{N.}},
\oauthor{\bsnm{Li}, \binits{Y.}},
\oauthor{\bsnm{Xia}, \binits{H.}}:
Deep learning segmentation model for automated detection of the opacity regions in the chest x-rays of the covid-19 positive patients and the application for disease severity.
medRxiv,
2020--10
(2020)
\end{botherref}
\endbibitem

\end{thebibliography}

\section*{Acknowledgments}
This research was partly supported by the National Natural Science Foundation of China (No. 62222118 to S. W. and No. U22A2040 to S.W.),
Guangdong Provincial Key Laboratory of Artificial Intelligence in Medical Image Analysis and Application (No. 2022B1212010011 to S.W.), 
Shenzhen Science and Technology Program (No. RCYX20210706092104034 to S.W. and No. JCYJ20220531100213029 to C.L.),
and Youth lnnovation Promotion Association CAS (S.W.).

\section*{Author contributions}
Weijian Huang: Conceptualization, methodology development, experiment, formal analysis, investigation, writing. 
Cheng Li: Formal analysis, investigation, validation, visualization, writing. 
Hong-Yu Zhou: Formal analysis, review and editing. 
Hao Yang: Methodology, investigation. 
Jiarun Liu: Methodology, investigation. 
Yong Liang: Review and editing. 
Hairong Zheng: Review and editing. 
Shaoting Zhang: Discussion, review and editing. 
Shanshan Wang: Conceptualization, methodology development, funding support, investigation, supervision, review and editing.

\section*{Competing interests}
The authors declare no competing interests.

\section*{Tables}

\begin{table}[h!]
\centering
\caption{Comparison of AUC scores (\%) for classification performance on three open-source datasets with varying ratios of annotated samples. Methods with and without zero-shot capabilities have both been included for comprehensive evaluation}
\setlength{\tabcolsep}{3.0pt}
\label{ftcls}
\begin{tabular}{ccccccccccc}
\hline
\multirow{2}{*}{\textbf{Method}} &
  \multirow{2}{*}{\textbf{Zero-shot}} &
  \multicolumn{3}{c}{\textbf{CheXpert}} &
  \multicolumn{3}{c}{\textbf{NIH}} &
  \multicolumn{3}{c}{\textbf{RSNA}}  \\ 
 & &
  \textbf{1\%} &
  \textbf{10\%} &
  \textbf{100\%} &
  \textbf{1\%} &
  \textbf{10\%} &
  \textbf{100\%} &
  \textbf{1\%} &
  \textbf{10\%} &
  \textbf{100\%} \\ \hline
  Ark  & $\times$    & - & - & 88.7 & - & - & 82.9 & - & - & 74.7 \\
M3AE    & $\times$   & 86.2 & 87.3 & 87.9 & -    & -    & -    & 89.0   & 90.8 & 92.3 \\
REFERS  & $\times$    & 87.2 & 88.1 & 88.2 & 76.7 & 80.9 & 84.7 & 89.4 & 91.6 & 92.7 \\
MRM     & $\times$ & 88.5 &  88.5 & 88.7 & 79.4 & 84.0 & 85.9 & 91.3 & 92.7 & 93.3 \\ \hline
ConVIRT & \checkmark & 85.9 & 86.8 & 87.3 & 66.2    & 76.6    & 81.3    & 77.4 & 80.1 & 81.3 \\
GLoRIA  & \checkmark & 86.6 & 87.8 & 88.1 & 67.1    & 76.6    & 81.3    & 86.1 & 88.0   & 88.6 \\
BioViL  & \checkmark & -    & -    & -    & 69.5    & 75.3    & 82.5    & 88.1 & 88.4 & 89.1 \\
MedKLIP & \checkmark & -    & -    & -    & 77.2 & 78.9 & 83.2 & 87.3 & 88.0   & 89.3 \\
M-FLAG   & \checkmark & -    & -    & -    & 62.2 & 71.6 & 78.7 & -    & -    & -    \\
MaCo (Ours) &
  \checkmark & 88.7&  88.7&  88.9 &  79.3 &  83.8 & 85.9 &  91.5 &  92.7 & 93.6 \\ \hline
\end{tabular}
\end{table}

\begin{table}[h!]
\centering
\caption{Disease-level classification performance (AUC: \%) on the NIH Chest X-ray dataset. Methods with and without zero-shot capabilities have both been included for comprehensive evaluation.}
\setlength{\tabcolsep}{2.0pt}
\label{nihftcls}

\begin{tabular}[h!]{ccccccccccccccccc}
\hline
Method &
Zero-shot &  Ate. &  Car. &  Eff. &  Inf. &  Mas. &  Nod. &  Pna. &  Pnx. &  Con. &  Ede. &  Emp. &  Fib. &  Thi. &  Her. &  AVG \\ \hline 
Model Genesis       & $\times$ & 78.8 & 84.5 & 86.6 & 71.1          & 81.9 & 73.2 & 73.0 & 85.6 & 79.2          & 87.8          & 86.6 & 81.0 & 75.8 & 85.2          & 81.0 \\
C2L                 & $\times$ & 81.1 & 90.2 & 88.0 & 72.0          & 82.7 & 74.1 & 75.3 & 85.9 & 81.0          & 88.1          & 88.0 & 80.8 & 76.2 & 86.8          & 82.2 \\
Context Restoration & $\times$ & 75.8 & 82.9 & 84.8 & 70.0          & 79.6 & 69.5 & 69.4 & 84.0 & 76.4          & 86.6          & 84.8 & 78.6 & 73.2 & 83.0          & 78.7 \\
TransVW             & $\times$ & 79.8 & 85.0 & 87.1 & 72.3          & 82.6 & 74.4 & 74.0 & 86.1 & 80.0          & 88.2          & 87.1 & 81.8 & 76.6 & 85.9          & 81.7 \\
REFERS              & $\times$ & 83.0 & 92.3 & 88.7 & 74.1 & 85.5 & 76.7 & 77.0 & 89.1 & 82.1          & 90.2          & 91.4 & 83.9 & 78.5 & 93.3          & 84.7 \\
MRM & $\times$ &  84.2 &  93.0 &  89.6 &  71.8 &  88.2 &  78.5 &  77.3 &  90.2 &  82.2 &  91.0 &  94.3 &  86.7 &  81.4 &  94.4 &  85.9 \\ \hline
ConVIRT             & \checkmark & 80.1 & 83.6 & 85.1 & 66.1          & 80.0 & 74.9 & 70.0 & 86.7 & 80.8          & 90.2          & 90.1 & 79.3 & 74.7 & 96.3          & 81.3 \\
GLoRIA              & \checkmark & 82.6 & 83.3 & 86.0 & 66.4          & 81.8 & 73.5 & 71.0 & 84.5 & 81.3          & 89.8          & 93.1 & 78.9 & 76.1 & 97.5          & 81.8 \\
BioViL              & \checkmark & 81.9 & 85.4 & 86.1 & 66.6          & 83.0 & 76.3 & 70.9 & 86.0 & 82.9          & 90.3          & 92.5 & 79.1 & 76.4 & 97.0          & 82.5 \\
MedKLIP             & \checkmark & 82.9 & 85.9 & 87.2 & 65.7          & 83.8 & 76.5 & 73.8 & 88.1 & 82.8 & 90.8 & 92.2 & 79.8 & 77.8 & 98.0 & 83.2 \\
MaCo (Ours) &  \checkmark &  84.3 &  92.8 &  89.4 &  72.4 &  87.8 &  78.8 &  77.5 &  90.0 &  82.2 &  90.7 &  94.4 &  87.0 &  80.3 &  95.2 &  85.9 \\ \hline
\end{tabular}
\end{table}

\begin{table}[h!]
\centering
\caption{Comparison of Dice scores (\%) for segmentation performance on the SIIM and COVID Rural datasets with varying ratios of annotated samples}
\begin{tabular}{ccccc}
\hline
\multirow{2}{*}{\textbf{Method}} & \multicolumn{2}{c}{\textbf{SIIM}} & \multicolumn{2}{c}{\textbf{COVID Rural}} \\ 
         & \textbf{10\%} & \textbf{100\%} & \textbf{10\%}  & \textbf{100\%} \\ \hline
MGCA     & 59.3          & 64.2           & -              & -              \\
M-FLAG   & 61.2          & 64.8           & -              & -              \\
Med-UniC & 62.2          & 64.4           & -              & -              \\
LoVT     & -             & 44.1           & -              & 51.2           \\
ConVIRT  & 43.2          & 59.9           & 27.2           & 37.4           \\
GLoRIA   & 46.9          & 63.4           & 28.1           & 38.7           \\
BioViL   & 62.7          & 70.0           & 32.4           & 41.6           \\
MedKLIP  & 72.1          & 79.4           & 35.4           & 44.0           \\
MaCo (Ours)     & 72.6 & 89.4  & 68.3 & 75.1  \\ \hline
\end{tabular}
\label{ftseg}
\end{table}

\begin{table}[h!]
\centering
\caption{Comparison of mean Average Precision (mAP) scores (\%) for detection performance on the RSNA dataset with varying ratios of annotated samples}
\begin{tabular}{cccc}
\hline
\multirow{2}{*}{\textbf{Method}} & \multirow{2}{*}{\textbf{Backbone}} & \multicolumn{2}{c}{\textbf{RSNA}} \\ 
         &        & \textbf{10\%} & \textbf{100\%} \\ \hline
ImageNet & ResNet & 12.4          & 8.0            \\
BYOL     & ResNet & 11.0          & 17.3           \\
SimCLR   & ResNet & 12.2          & 18.8           \\
PixelPro & ResNet & 11.0          & 17.4           \\
CLIP     & ResNet & 10.7          & 19.9  \\
LoVT     & ResNet & 13.2 & 18.1           \\ \hline
CLIP*    & VIT    & 10.6          & 17.4           \\
MaCo (Ours)     & VIT    & 11.9 & 19.2  \\ \hline
\end{tabular}
\label{ftdetection}
\end{table}

\begin{table}[h!]
\centering
\caption{Comparison of AUC scores (\%) for zero-shot classification on the NIH, RSNA, and SIIM datasets.}
\begin{tabular}{cccc}
\hline
\textbf{Method} & \textbf{NIH}  & \textbf{RSNA} & \textbf{SIIM} \\ \hline
ConVIRT         & 61.0          & 80.4          & 64.3          \\
GLoRIA          & 66.1          & 71.5          & 53.4          \\
BioViL          & 69.1          & 82.8          & 70.8          \\
CheXzero        & 73.0          & 85.8          & 68.8          \\
MedKLIP        &  76.8          & 86.9          & 89.2          \\
MaCo (Ours)            & 77.3 & 88.6 & 90.4 \\ \hline
\end{tabular}
\label{zscls}
\end{table}

\begin{table}[h!]
\centering
\caption{Comparison of CNR and mIoU for zero-shot phrase grounding on the MS-CXR datasets.}
\begin{tabular}{cccc}
\hline
\textbf{Method} & \textbf{Pretrained dataset}    & \textbf{CNR}            & \textbf{mIoU (\%)}        \\ \hline
BioViL  & PubMed + MIMIC-III + MIMIC-CXR & 1.027 & 26.6 \\ \hline
ConVIRT & MIMIC-CXR                      & 0.818          & 23.8          \\
GLoRIA  & MIMIC-CXR                      & 0.930          & 24.6          \\
MaCo (Ours)    & MIMIC-CXR                      & 1.144 & 25.5 \\ \hline
\end{tabular}
\label{pg}
\end{table}

\begin{table}[h!]
\setlength{\tabcolsep}{2.pt}
\centering
\caption{Ablation study results with zero-shot phrase grounding and fine-tuning classification experiments on the MS-CXR and RSNA datasets}
\begin{tabular}{ccccccc}
\hline
\multirow{2}{*}{\textbf{Method}} & \multicolumn{3}{c}{\textbf{MS-CXR}} & \multicolumn{3}{c}{\textbf{RSNA}} \\ 
                         & \textbf{CNR}   & \textbf{mIoU}       & \textbf{PG}     & \textbf{1\%} & \textbf{10\%} & \textbf{100\%} \\ \hline
                         
MAE                      & -       & -     & -  & 83.2  & 89.2   & 91.0    \\
+HR                      & -       & -     & -  & 83.3  & 89.3   & 91.0    \\
+CLIP                    & 0.860   & 21.2      & 0.330  & 90.9  & 92.1   & 93.0    \\
+Correlation Weighting    & 1.144   & 25.5      & 0.373  & 91.5  & 92.7   & 93.6    \\ \hline
\end{tabular}
\label{ablation}
\end{table}

\begin{table}[h]
\renewcommand{\tablename}{Supplementary Table.}
\caption{The effect of the soft threshold $\tau^{w}$ in the softmax function on phrase-grounding performance.}
\begin{tabular}{ccc}
\hline
\textbf{$\tau^{w}$} & \textbf{CNR}   & \textbf{mIoU (\%)} \\ \hline
0.05         & 1.109          & 25.1          \\
0.04         & 1.118          & 25.2          \\
0.03         & 1.129          & 25.5          \\
0.02         & 1.144          & 25.5 \\
0.01         & 1.149 & 24.8         \\ 
0.005        & 1.095 & 22.4        \\\hline
\end{tabular}
\label{tau}
\end{table}

\section*{Figure}

\begin{figure*}[htb]
\centering
\includegraphics[scale=0.4]{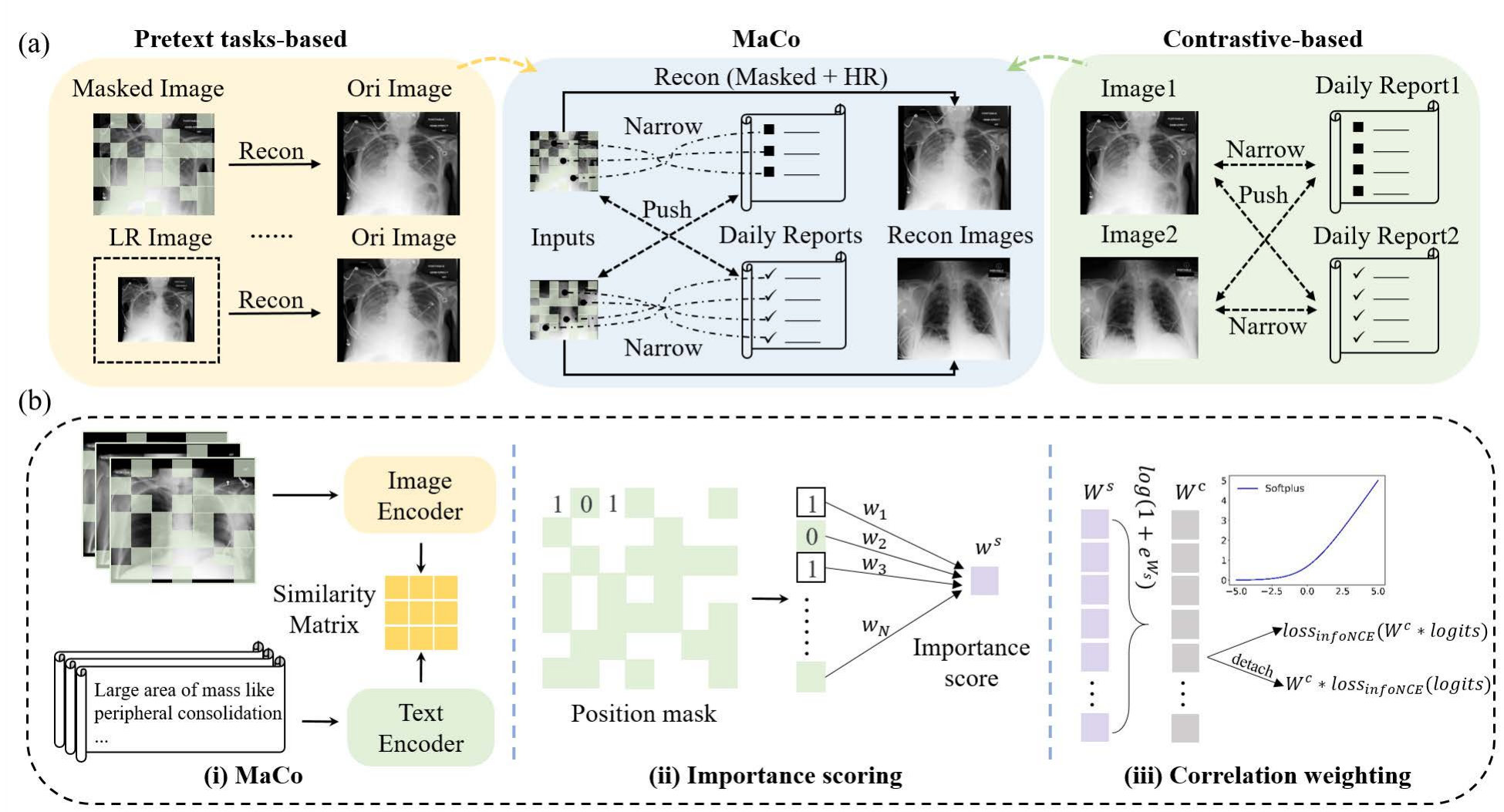}
\caption{The proposed MaCo framework. (a) An illustration of the masked contrastive learning strategy employed in MaCo, which leverages the advantages of both contrastive learning and pretext tasks. LR denotes the low-resolution image obtained after downsampling, while HR refers to the original high-resolution image. (b) The proposed correlation weighting mechanism, (i) shows the basic structure of MaCo, where image and text representations are compared using a contrastive loss, (ii) presents the procedure to generate the importance score, and (iii) plots the method to build correlations.}
\end{figure*}

\begin{figure*}[htb]
\centering
\includegraphics[scale=0.6]{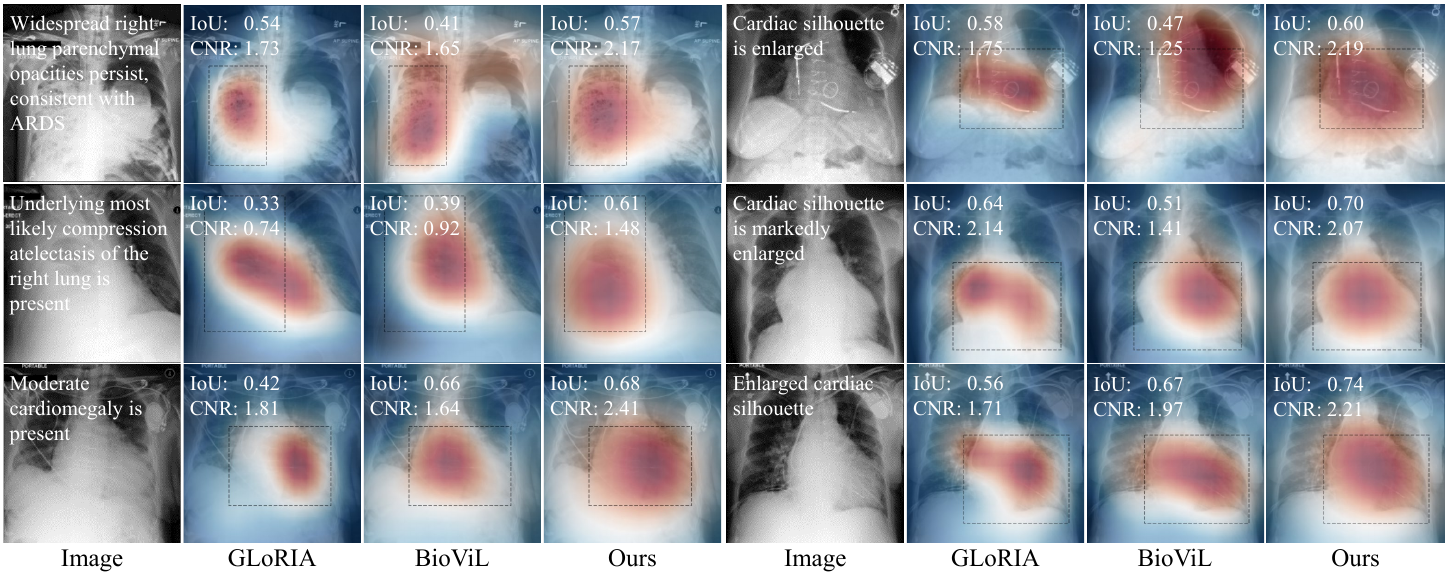}
\caption{Qualitative phrase-grounding results when provided with description phrases. We visualize the association of vision and language on the MS-CXR dataset. The description phrases are marked in white font in the image column. The gold standard annotations outlined by clinical experts are represented with dashed boxes.}
\end{figure*}

\begin{figure*}[htb]
\centering
\includegraphics[scale=0.4]{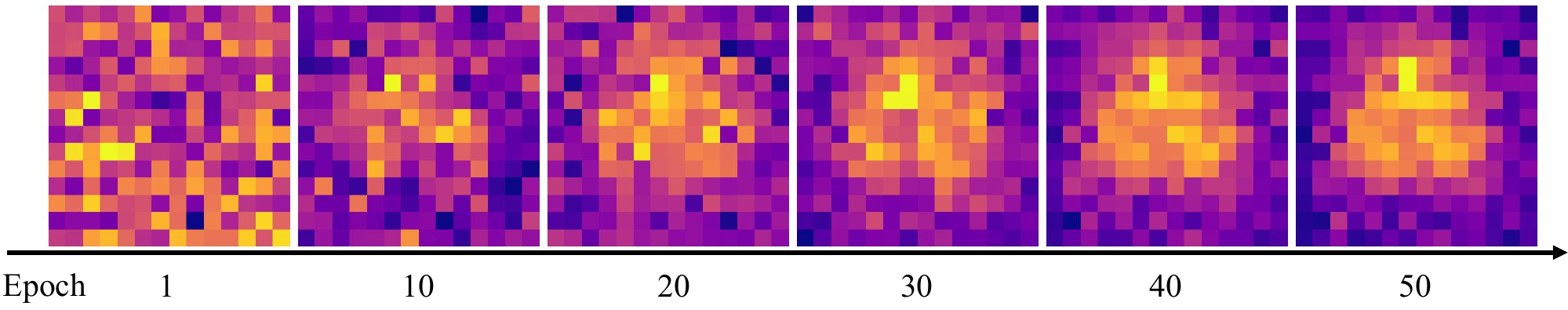}
\caption{Visualization of the weights of the proposed correlation weighting mechanism. The number under the picture indicates the training epoch. After training, the weights are larger in the central regions with a higher incidence of disease and smaller in the background regions around the edges.}
\end{figure*}

\begin{figure*}[htb]
\centering
\renewcommand{\figurename}{Supplementary Fig.}
\includegraphics[scale=0.4]{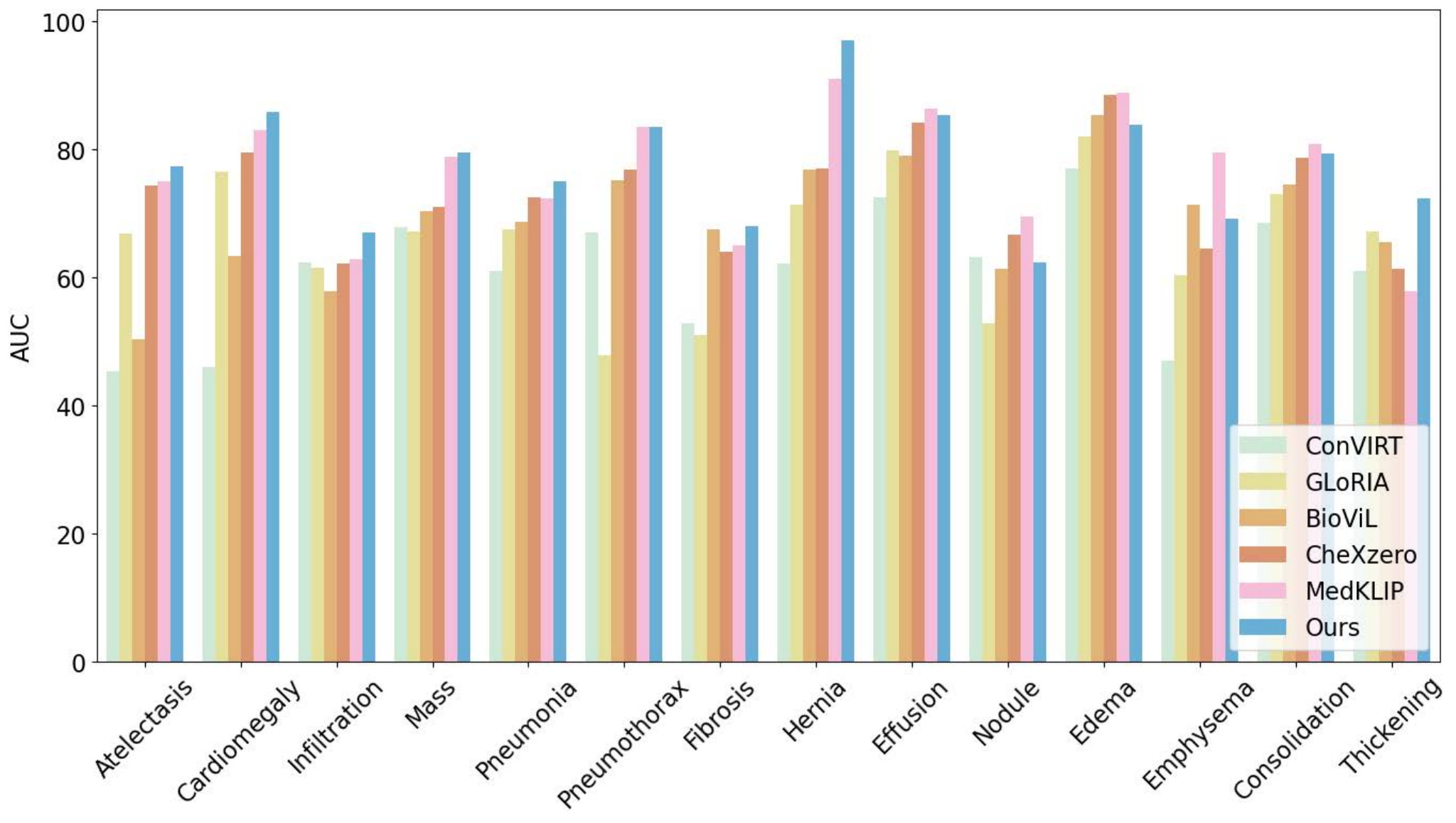}
\caption{Disease-level zero-shot classification performance of different methods on the NIH Chest X-ray dataset}
\end{figure*}

\end{document}